\documentclass{article}

\usepackage[T1]{fontenc}
\usepackage{microtype}
\usepackage{graphicx}
\graphicspath{{../}}
\usepackage{subcaption}
\usepackage{booktabs} % for professional tables

\usepackage{hyperref}

\usepackage[accepted]{arxiv}

\usepackage{amsmath}
\usepackage{amssymb}
\usepackage{mathtools}
\usepackage{amsthm}
\usepackage{url}
\usepackage{algorithm}
\usepackage{algorithmic}
\usepackage{booktabs}
\usepackage{graphicx}
\usepackage{svg}
\usepackage{enumerate}
\usepackage{comment}
\usepackage{xcolor}
\usepackage{colortbl}
\usepackage{tablefootnote}
\usepackage{makecell}
\usepackage{pifont}
\usepackage{bbding}
\usepackage{accents}
\usepackage{graphicx}
\usepackage{thmtools}
\usepackage{thm-restate}

\usepackage{wrapfig}
\allowdisplaybreaks
\usepackage[capitalize,noabbrev]{cleveref}

\theoremstyle{plain}

\theoremstyle{definition}

\theoremstyle{remark}

\crefname{assumption}{Assumption}{Assumptions}

\newcommand{\crefdefpart}[2]{%
  \hyperref[#2]{\namecref{#1}~\labelcref*{#1}(\ref*{#2})}%
}

\newcommand{\refdefpart}[2]{%
  \hyperref[#2]{\labelcref*{#1}(\ref*{#2})}%
}

\newcommand{\factdefpart}[2]{%
  \hyperref[#2]{Fact~(\ref*{#2})}%
}

\newcommand{\halfcheck}{\ding{51}\textsuperscript{\kern-0.55em\ding{55}}}

\usepackage{amsmath,amsfonts,bm}

\def\eqref#1{equation~\ref{#1}}
\def\1{\bm{1}}

\DeclareMathAlphabet{\mathsfit}{\encodingdefault}{\sfdefault}{m}{sl}
\SetMathAlphabet{\mathsfit}{bold}{\encodingdefault}{\sfdefault}{bx}{n}

\DeclareMathOperator*{\argmin}{arg\,min}

\usepackage[textsize=tiny]{todonotes}

\icmltitlerunning{BLISS: A Lightweight Bilevel Influence Scoring Method for Data Selection in Language Model Pretraining}

\begin{document}

\twocolumn[
  \icmltitle{BLISS: A Lightweight Bilevel Influence Scoring Method for Data Selection in Language Model Pretraining}

  % It is OKAY to include author information, even for blind submissions: the
  % style file will automatically remove it for you unless you've provided
  % the [accepted] option to the icml2026 package.

  % List of affiliations: The first argument should be a (short) identifier you
  % will use later to specify author affiliations Academic affiliations
  % should list Department, University, City, Region, Country Industry
  % affiliations should list Company, City, Region, Country

  % You can specify symbols, otherwise they are numbered in order. Ideally, you
  % should not use this facility. Affiliations will be numbered in order of
  % appearance and this is the preferred way.
  \icmlsetsymbol{equal}{*}

  \begin{icmlauthorlist}
    \icmlauthor{Jie Hao}{GMU}
    \icmlauthor{Rui Yu}{GMU}
    \icmlauthor{Wei Zhang}{IBM}
    \icmlauthor{Huixia Judy Wang}{RU}
    \icmlauthor{Jie Xu}{GMUS}
    \icmlauthor{Mingrui Liu}{GMU}
    %\icmlauthor{}{sch}
  \end{icmlauthorlist}

  \icmlaffiliation{GMU}{Department of Computer Science, George Mason University, USA,}
  \icmlaffiliation{GMUS}{Department of System Engineering \& Operations Research, George Mason University, USA}
  \icmlaffiliation{IBM}{IBM T.J. Watson Research Center, USA, }
  \icmlaffiliation{RU}{Department of Statistics, Rice University,}

  \icmlcorrespondingauthor{Mingrui Liu}{mingruil@gmu.edu}

  % You may provide any keywords that you find helpful for describing your
  % paper; these are used to populate the "keywords" metadata in the PDF but
  % will not be shown in the document
  \icmlkeywords{Machine Learning, ICML}

  \vskip 0.3in
]

% this must go after the closing bracket ] following \twocolumn[ ...

% This command actually creates the footnote in the first column listing the
% affiliations and the copyright notice. The command takes one argument, which
% is text to display at the start of the footnote. The \icmlEqualContribution
% command is standard text for equal contribution. Remove it (just {}) if you
% do not need this facility.

% Use ONE of the following lines. DO NOT remove the command.
% If you have no special notice, KEEP empty braces:
\printAffiliationsAndNotice{}  % no special notice (required even if empty)
% Or, if applicable, use the standard equal contribution text:
% \printAffiliationsAndNotice{\icmlEqualContribution}

\begin{abstract}

Effective data selection is essential for pretraining large language models (LLMs), improving efficiency and generalization to downstream tasks. However, existing approaches often rely on external pretrained models, making it difficult to separate the benefits of data selection from those introduced by external models. In addition, many methods estimate data importance from a fixed model state or short-horizon update, making it hard to capture how data preference changes as the model evolves during pretraining. In this paper, we introduce BLISS (\textbf{B}ileve\textbf{L} \textbf{I}nfluence \textbf{S}coring method for data \textbf{S}election), a lightweight data selection method that operates entirely \emph{from scratch}, without external pretrained oracle models, while modeling dynamic data preference. BLISS uses a small proxy model as a surrogate for the LLM and trains a score model to estimate sample importance through multi-step proxy updates induced by score-weighted training data. We formulate data selection as a bilevel optimization problem: the upper-level objective optimizes the score model to assign sample weights, so minimizing the lower-level weighted training loss improves validation performance. Once optimized, the score model predicts influence scores, enabling efficient selection of high-quality samples for LLM pretraining. We validate BLISS by pretraining 410M/1B/2.8B Pythia and LLaMA-0.5B models on selected C4 subsets. Under the 1B setting, BLISS achieves a $1.7\times$ speedup in reaching the same performance as the state-of-the-art method, while delivering superior performance across multiple downstream tasks.

\end{abstract}

\section{Introduction}

The successful large-scale language model pretraining crucially relies on the careful choice of pretraining data~\citep{brown2020language,raffel2020exploring,du2022glam,elazar2023s}. Recent studies have shown that effective data selection (a.k.a., data curation) methods can enhance pretraining efficiency~\citep{xie2023doremi} and improve generalization~\citep{engstrom2024dsdm,wettig2024qurating}. There are various types of data selection approaches for language model pretraining, including language filtering~\citep{laurenccon2022bigscience,wenzek2019ccnet},  data deduplication~\citep{lee2021deduplicating,abbas2023semdedup}, heuristic approaches~\citep{rae2021scaling,penedo2023refinedweb}, data-quality filtering~\citep{brown2020language,gao2020pile,chowdhery2023palm,xie2023data,wettig2024qurating}, data mixing~\citep{xie2023doremi,albalak2023efficient,xia2023sheared}, and data influence function based methods~\citep{park2023trak,engstrom2024dsdm,yu2024mates}. Despite the rich literature of data selection methods in large language model (LLM) pretraining (e.g., a survey paper in~\citet{albalak2024survey}), it is still unclear what properties are needed for the training data curation to guarantee good performance: it remains an important real-world challenge~\citep{li2024datacomp}. Beyond natural language, efficient generative modeling and data curation are also becoming increasingly important in scientific domains, such as geoscience or on-device data generation~\citep{yang2025novel,yang2026digit}.

Existing approaches of data selection methods suffer from two major limitations. First, they often require leveraging pretrained models~\citep{brown2020language,xie2023data,wettig2024qurating} for data-quality filtering, making it difficult to separate the effects of data selection from those of the external pretrained models. For example, the QuRating method~\citep{wettig2024qurating} assigns quality ratings to training samples based on responses from a pretrained LLM (e.g., GPT-3.5) before training a QuRater model. This reliance raises uncertainty about the role of the external LLM in the training process and whether its feedback is truly optimal. Moreover, the cost of invoking these external pretrained models is prohibitively expensive during data selection process for large-scale pretraining. Second, many influence-based approaches estimate data utility from a fixed model state or a short-horizon update. For example, the data influence function based approach~\citep{yu2024mates} evaluates the impact of individual training samples based on a single training step with the current model, which does not capture the cumulative effects of data selection over the course of full model training. 

\begin{figure}
    \centering
    \includegraphics[width=1\linewidth]{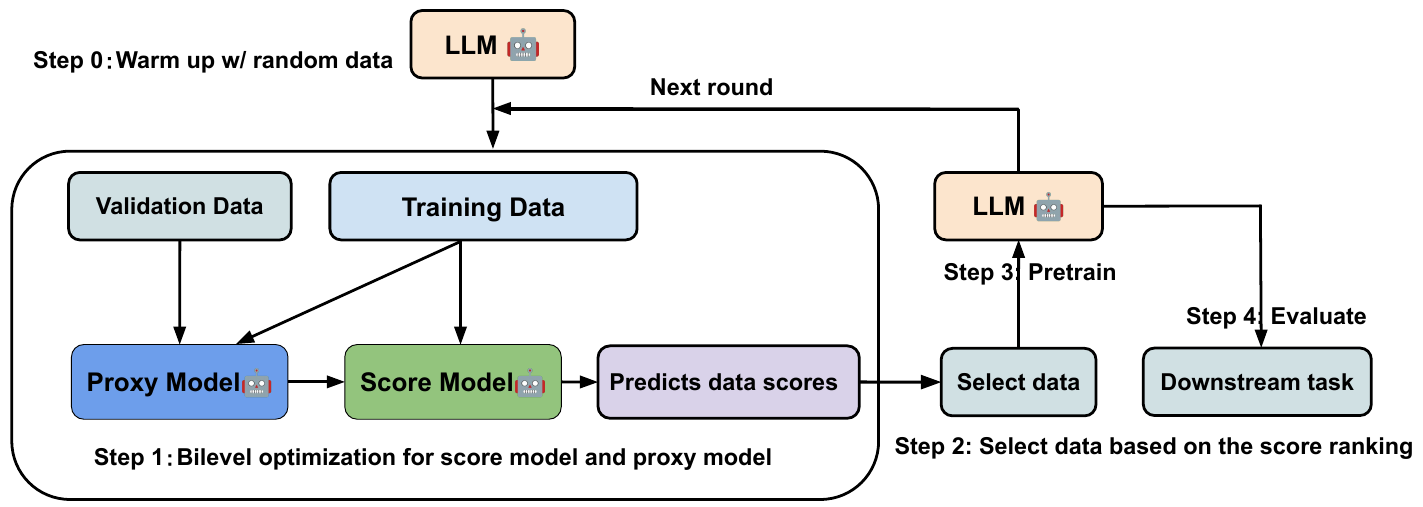}
    \caption{The pipeline of data selection and pretraining procedure. There are four main steps in one round training, 1) Warm up LLM using randomly selected training data (e.g. 10k step); 2) Bilevel optimization for score and proxy model, 3) Predict the data influence, and select Top-20\% training data based on their score ranking; 4) Retrain the LLM using the selected data (e.g., 10k steps); 5) Evaluate on the downstream task. Repeating the above steps can achieve multiple-round training.}
    \label{fig:illustration}
\end{figure}

In this paper, we introduce a new data selection method to address the two major limitations of existing approaches. Our method, namely BLISS (\textbf{B}ileve\textbf{L} \textbf{I}nfluence \textbf{S}coring method for data \textbf{S}election), is a lightweight data selection method that operates entirely \emph{from scratch}, without relying on any external pretrained models, while explicitly modeling dynamic data preference under model updates. The core innovation of our approach lies in \emph{the integration of two lightweight models within a novel bilevel optimization framework} for data selection. Our method bypasses traditional data-quality filtering and uses bilevel optimization
to adapt sample weights according to the evolving proxy-model training dynamics. In particular, BLISS leverages a small proxy model as a surrogate for the LLM and employs a score model to estimate sample importance through score-induced proxy-model updates. Our bilevel optimization problem has upper-level and lower-level objectives: the upper-level objective optimizes the score model to assign importance weights to training samples, ensuring that minimizing the lower-level objective (i.e., training the proxy model over the weighted training loss until convergence) leads to best validation performance. Once the bilevel optimization is solved, the trained score model predicts influence scores for the entire dataset, enabling the selection of high-score samples for LLM pretraining. The pipeline of our proposed procedure is illustrated in Figure~\ref{fig:illustration}. The main contributions of our paper are summarized as the following:

% \vspace*{-0.05in}
\begin{itemize}
\item We propose a principled approach to data selection for language model pretraining. Our method, BLISS, leverages a novel bilevel optimization framework that employs a proxy model and a score model to model dynamic data preference under model updates. Unlike existing methods, BLISS operates from scratch without relying on any pretrained oracle models for data-quality filtering, obviating any biases or risks that may arise from such dependence\footnote{Many commercial large-scale pretrained models strictly prohibit users from generating data or using them to facilitate the training of other models, as doing so may result in severe legal consequences~\citep{openai2024terms, google2024geminiterms}. Our approach is entirely free from such legal concerns. We rely solely on algorithmic advancements applied to a model trained from scratch, without any dependence on third-party pretrained large-scale models.}. 

\item We validate our method by pretraining 410M/1B Pythia and LLaMA-0.5B models on selected subsets of C4 dataset. Experimental results on 1B setting demonstrate a $1.7\times$ speedup in reaching the same performance as the state-of-the-art method such as MATES~\citep{yu2024mates}. Furthermore, we scale up our experiments to pretraining a 2.8B model using the data selected in the 1B experiment, and demonstrate that our method consistently outperforms MATES at every round of data selection, achieving $1.4\%$ performance improvement over MATES~\citep{yu2024mates}.

\item Through extensive ablation studies, we demonstrate the effectiveness of each component in our bilevel optimization framework, further substantiating the robustness and efficiency of our approach.

\end{itemize}

\section{Related Work}

\textbf{Data Selection for Language Model Training.} Early approaches to data selection primarily relied on rule-based methods as language filters for training data, employing utility functions tailored to specific datasets~\citep{conneau2019cross,raffel2020exploring,rae2021scaling,penedo2023refinedweb}. Another key category is data deduplication~\citep{lee2021deduplicating,sorscher2022beyond,penedo2023refinedweb,abbas2023semdedup,tirumala2023d4}, which eliminates redundant samples to optimize training efficiency and enhance performance on downstream tasks. A class of methods exist for performing data-quality filtering, which can select data similar to high-quality corpus of data points~\citep{brown2020language,du2022glam,gao2020pile,xie2023data,li2024datacomp}, with small perplexity~\citep{chowdhery2023palm,wenzek2019ccnet}. More recent methods leverage external pretrained LLMs to evaluate the pretraining data quality~\citep{wettig2024qurating,maini2024rephrasing,zhuang2025meta}. In addition, a similar variant of data selection is domain reweighting for data mixtures~\citep{oren2019distributionally,sagawa2019distributionally,xie2023doremi,fan2023doge,albalak2023efficient,chen2023skill}, which re-scale the contribution of each domain to enhance generalization. Another recently emerged line of research leverages the tool of influence functions~\citep{hampel1974influence,cook1977detection,ling1984residuals,koh2017understanding} to evaluate the impact of individual training samples on a fixed LLM~\citep{park2023trak,engstrom2024dsdm,yu2024mates,pan2025alinfik,lin2024token}. QUAD \citep{zhang2409harnessing} proposes an efficient framework incorporating the attention layers to estimate the influence scores.
Different from token-level attribution methods such as \citep{lin2024token}, which retrieve influential training data for a given token prediction of an already trained or fixed LLM, BLISS targets sample-level data selection for pretraining from scratch. BLISS also differs from classical influence-estimation approaches such as
DataInf \citep{kwon2024datainf}, which approximate the effect of upweighting or removing individual examples around a fixed fine-tuning state. Moreover, methods such as ALinFiK \citep{pan2025alinfik} estimate per-sample influence through training/test gradient interactions on the target LLM. In contrast, BLISS does not compute influence scores directly on the target LLM; instead, it learns a scoring model through lightweight proxy-based bilevel optimization, with a KL alignment term encouraging the proxy model to reflect the target model's data preference.

In contrast to these works, our method models dynamic data preference induced by multi-step proxy-model updates, rather than estimating influence only around a fixed model state. In addition, our method can train the model from scratch and does not need any extra information from any external pretrained models, making it a scalable and effective solution.

% \vspace*{-0.05in}
\textbf{Bilevel Optimization and Data Selection.} Bilevel optimization provides a powerful framework for modeling optimization problems with a nested structure~\citep{bracken1973mathematical,dempe2002foundations}. Recent research has focused on developing efficient bilevel optimization algorithms with strong theoretical guarantees~\citep{ghadimi2018approximation,hong2023two,ji2021bilevel,kwon2023fully,dagreou2022framework,chen2023bilevel1,grazzi2022bilevel,hao2024bilevel,gong2024a,gong2024accelerated,gong2026adaptive,wu2026bilevel}. This approach has been widely applied in various machine learning tasks, including meta-learning~\citep{finn2017model}, hyperparameter optimization~\citep{franceschi2018bilevel}, and natural language processing~\citep{somayajula2023bi,grangier2023bilevel}. For the application of data selection, bilevel optimization has been utilized for continual learning~\citep{borsos2020coresets,zhou2022probabilistic,hao2023bilevel1} and data reweighting in LLM fine-tuning~\citep{pan2024scalebio,shen2024seal}. Our work is most closely related to SEAL \citep{shen2024seal}, which focuses on selecting high-quality and safe data to fine-tune a pretrained LLM, with the goal of aligning the model with safety and ethical guidelines. However, our approach differs from SEAL in two key aspects:
(1) Problem setting. While SEAL operates in a fine-tuning context, our objective is to select data for \textbf{pretraining} an LLM \textbf{from scratch}, aiming to improve downstream performance \textbf{without relying on any external pretrained models}.
(2) Model update mechanism. SEAL utilizes the LoRA technique~\citep{hu2022lora} to update both the data selector and the LLM during fine-tuning. However, this approach is not directly applicable to our setting due to the following reasons. First, LoRA is only suitable for fine-tuning tasks but insufficient for full model pretraining. Second, their algorithm always updates the original large models directly, which is computationally expensive if all parameters are updated. In contrast, we propose a more efficient framework that introduces lightweight models (a score model and a proxy model) to guide data selection, while allowing full parameter updates within these smaller networks.
To the best of our knowledge, our proposed bilevel influence scoring method is the first to leverage bilevel optimization techniques for data selection in LLM pretraining.

\section{Preliminaries and Notations} 

Suppose that we have a large-scale training dataset $\mathcal{D}_{tr} = \{\xi_{i} \mid 0\leq i\leq N-1\}$ and a downstream task $\mathcal{D}_{ds}$. The goal is to select a subset of training set, namely $\mathcal{D}_s\subset \mathcal{D}_{tr}$ and $|\mathcal{D}_s| = Q\leq N$, to  pretrain a large language model with a specific training budget (e.g., limited FLOPs), such that the model can achieve high performance on the downstream task $\mathcal{D}_{ds}$. Generally, the downstream data is inaccessible during pretraining. Instead, we can use a validation data $\mathcal{D}_{val}=\{\zeta_i\mid 0\leq i\leq M-1\}$ to estimate the model's performance on $\mathcal{D}_{ds}$, because these two datasets often have similar data distributions or share common domain knowledge. A small subset of training data $\tilde{\mathcal{D}}_{tr} \subset \mathcal{D}_{tr}$ is uniformly sampled from $\mathcal{D}_{tr}$.

In bilevel optimization, $f(\cdot)$ and $g(\cdot)$ denote the upper-level (UL) and lower-level (LL) functions, respectively. Machine learning often requires solving stochastic optimization problems $f(\cdot)=\mathbb{E}_{\xi\sim \mathcal{D}_f}[F(\cdot; \xi)]$ and  $g(\cdot)=\mathbb{E}_{\zeta\sim \mathcal{D}_g}[G(\cdot; \zeta)]$, where $\mathcal{D}_f$ and $\mathcal{D}_g$ are the underlying unknown data distribution for $f$ and $g$, respectively. $F(\cdot;\xi)$ denotes the upper-level stochastic objective function and $G(\cdot;\zeta)$ is the lower-level stochastic objective function. Noisy observations of $f$ and $g$ can be collected by sampling from $\mathcal{D}_f$ and $\mathcal{D}_g$. 

\section{Methods} 

\subsection{Bilevel Influence Scoring Framework} \label{sec:bliss_frame}
The goal of data selection is to optimize the performance of the LLM on downstream tasks by training it using an optimal subset of training data. However, directly searching for the optimal subset of training samples faces prohibitive costs due to the combinatorial nature of the problem and the high computational cost of estimating the performance of the LLM for every potential subset being evaluated.

To address the aforementioned computational challenge, our bilevel influence scoring framework uses a lightweight score model $\theta_s$ to predict the influence of every sample on the model's performance for the downstream task. The optimized score model is then used to infer the influence score of training samples, enabling the selection of the subset with the highest influence, thus streamlining the process to search for the optimal training data. Instead of directly estimating the performance of LLM (parameterized by $\theta_{tr}$) which is computationally expensive, our framework introduces a lightweight proxy model $\theta_{p}$ to approximate the behavior of the LLM. Note that the score model and the proxy model are both small models: they share a similar architecture and number of parameters. To ensure the data preferences of the proxy model align with those of the LLM, we apply knowledge distillation by minimizing the Kullback-Leibler (KL) divergence between the output logits of the proxy model and the LLM. We formulate the bilevel optimization for data selection as follows:
% \begin{equation}\label{eq:obj}
% \begin{aligned}
%     \min_{\theta_{s}} \Phi(\theta_s)
%     &\coloneqq f(\theta_p^*(\theta_s))
%       \coloneqq \mathbb{E}_{\zeta\sim \mathcal{D}_{val}}
%       F(\theta_p^*(\theta_s); \zeta)
%       \qquad \ \text{(UL)},\\[0.2em]
%     \text{s.t. } \qquad
%    & \theta_p^*(\theta_s)
% = \argmin_{\theta_p} g(\theta_p,\theta_s)
% \qquad \text{(LL)},
% \end{aligned}
% % \vspace*{-0.15in}
% \end{equation}

\begin{align}
  \min_{\theta_{s}} \Phi(\theta_s)
  &\coloneqq f(\theta_p^*(\theta_s))
    \coloneqq \mathbb{E}_{\zeta\sim \mathcal{D}_{val}}
    F(\theta_p^*(\theta_s); \zeta)
    \qquad \text{(UL)}, \nonumber \\[0.2em]
  &\text{s.t. } \quad
 \theta_p^*(\theta_s)
= \argmin_{\theta_p} g(\theta_p,\theta_s)
\quad \text{(LL)}, \label{eq:obj}
\end{align}

\noindent where the lower-level objective is defined as
\[
\begin{aligned}
g(\theta_p,\theta_s)
&=
\sum_{i=1}^{N} P_i
\mathcal{L}(\theta_p;\xi_i)
+
\gamma D_{KL}\big(
\ell(\theta_{tr};\xi_i)\|
\ell(\theta_p;\xi_i)
\big) \\
&\quad+
\lambda\|\theta_p\|^2 .
\end{aligned}
\]
Here, $f$ denotes the loss function on the valdation data $\mathcal{D}_{val}$
$P_i = \frac{e^{h(\theta_s; \xi_i)}}{\sum_{j=1}^{N} e^{h(\theta_s; \xi_j)}}$
represents the normalized importance weight of a data $\xi_i$ sampled from training data $\mathcal{D}_{tr}$, and
$h(\theta_s;\cdot): \mathbb{R}^{d_x}\rightarrow \mathbb{R}$ maps a sample to an unnormalized influence score.
$\mathcal{L}(\cdot)$ and $F(\cdot)$ denote the loss functions for next-token prediction, with a common choice being cross-entropy.
The softmax-normalized output distributions are represented by $\ell(\cdot)$.
The KL divergence is defined as $D_{KL}(X \| Y) = \sum_i X_i \log(\frac{X_i}{Y_i})$.
$\gamma$ and $\lambda$ are the regularization coefficients for the KL-divergence and weight-decay terms, respectively.

\textbf{Intuition of Algorithm Design.}
The lower level asks: \emph{given a candidate weighting over
training samples, what proxy model will be obtained after training under these
weights?} It updates the proxy model on the score-weighted training loss,
together with a KL alignment term that encourages the proxy to reflect the data
preference of the target LLM. The upper level then asks: \emph{do these weights
lead to better validation performance?} It updates the score model so that the
resulting proxy model $\theta_p^*(\theta_s)$ achieves lower validation loss.
In this way, the score model proposes sample weights, the proxy model reflects
their effect through multi-step training dynamics, and the validation loss
provides feedback to improve the score model.

This differs from one-step influence methods such as MATES~\citep{yu2024mates},
which estimate sample utility from a local update at the current model state.
BLISS instead learns sample weights through a bilevel interaction between the
score model and a dynamically updated proxy model, allowing the selected data to
better reflect the evolving model state. Moreover, the framework in
\eqref{eq:obj} does not rely on external pretrained oracle models, making BLISS
a self-contained data selection approach for pretraining from scratch.

\subsection{Algorithm for Updating the Proxy Model and Score Model} \label{sec:algorithm}

Now we design efficient algorithms for solving the bilevel problem (\ref{eq:obj}). The lower-level problem aims to optimize the proxy model $\theta_p$ on the weighted training samples with the influence predicted by the score model. Note that we freeze the LLM ($\theta_{tr}$) through the process of solving the bilevel optimization problem, as the LLM is used to infer the output logits. Therefore,  we perform the following update for the lower-level objective on a mini-batch of size $\mathcal{B}$ (full batch is infeasible in practice):
%\vspace*{-0.1in}
% \begin{equation}
%     \begin{aligned}
%         \theta_p^{t+1} 
%         &= \theta_p^{t} - \eta_1 \nabla_{ \theta_p} \sum_{i=1}^{\mathcal{B}}G(\theta_p^{t}, \theta_s^{t};\xi_i) \\
%         &= \theta_p^{t} - \eta_1\sum_{i=1}^{\mathcal{B}} \Bigg( P_i \nabla_{\theta_p} \mathcal{L}(\theta_p^t; \xi_i) \\
%         & \quad+ \gamma \sum_j \nabla_{\theta_p} \ell_{j}(\theta_p^t; \xi_i) 
%         \log \frac{\ell_j(\theta_p^t; \xi_i)}{\ell_j(\theta_{tr}^t; \xi_i)} 
%         + 2\lambda \theta_{p}^t \Bigg),
%     \end{aligned}
% \end{equation}

\begin{align}
\theta_p^{t+1} 
&= \theta_p^{t} - \eta_1 \nabla_{ \theta_p} \sum_{i=1}^{\mathcal{B}}G(\theta_p^{t}, \theta_s^{t};\xi_i)
\nonumber \\
&= \theta_p^{t} - \eta_1\sum_{i=1}^{\mathcal{B}} \Bigg(
P_i \nabla_{\theta_p} \mathcal{L}(\theta_p^t; \xi_i) \\
&\quad+ \gamma \sum_j \nabla_{\theta_p} \ell_{j}(\theta_p^t; \xi_i) 
\log \frac{\ell_j(\theta_p^t; \xi_i)}{\ell_j(\theta_{tr}; \xi_i)} \Bigg)
- 2\eta_1\lambda \theta_{p}^t,
 \nonumber
\end{align}

where $\ell_j(\cdot)$ denotes the $j$-th normalized logit of the output. Note that the exact computation of $P_i$ depends on all $N$ samples, which is computationally infeasible. Therefore, we approximate $P_i$ by replacing the full summation in the denominator with a partial summation over a smaller subset. This approximation is implemented in a distributed manner, significantly reducing the computational overhead. More details can be found in Appendix~\ref{sec:distributedsoftmax}.
For the upper-level update, we take the derivative of $\Phi(\theta_s)$ with respect to $\theta_s$ by chain rule, which is known as the hypergradient:
\begin{equation}
\label{eq:stochastichyper}
    \begin{aligned}
        &\nabla_{\theta_s}\Phi(\theta_s)\\
        &= -\nabla_{\theta_s\theta_p}^2g(\theta_p^*(\theta_s), \theta_s)\underbrace{[\nabla_{\theta_p}^2g(\theta_p^*(\theta_s), \theta_s)]^{-1}\nabla_{\theta_p}f(\theta_p^*(\theta_s))}_{z}, 
    \end{aligned}
   %\vspace*{-0.03in}
\end{equation}
% \vspace*{-0.01in}
where $z$ is the solution of the quadratic function $\min_z \frac{1}{2} z^T 
\nabla_{\theta_p}^2g(\theta_p^*(\theta_s), \theta_s) z - z^T\nabla_{\theta_p} f(\theta_p^*(\theta_s)) $. It can be solved by running a few steps of gradient descent in practice:
\begin{equation}\label{eq:z_update}
    z_{k+1}^t = z_k^t - \eta_2 \big( \nabla_{\theta_p}^2g(\theta_p^t, \theta_s^t) z_k^t - \nabla_{\theta_p} f(\theta_p^t)  \big),
    %\vspace*{-0.05in}
\end{equation}
where $k$ is the number of gradient updates for updating $z$ at a fixed iteration $t$ of updating $\theta_s$. We run 3 steps of gradient descent to solve $z$ in our experiments. Note that Equation (\ref{eq:z_update}) computes the Hessian-Vector-Product (HVP) term $\nabla_{\theta_p}^2g(\theta_p^t, \theta_s^t) z_k^t$ and thus avoids the computationally prohibitive operation of taking the inverse of the Hessian. The dimension of $z$ is the same as that of the parameters of the lightweight proxy model. Therefore, the computation of HVP within the PyTorch framework is quite similar to that of gradient. In our implementation, we use the stochastic variants of~\Cref{eq:stochastichyper} and~\Cref{eq:z_update} for updating the score model. In particular, the approximation of hypergradient at iteration $t$ on the mini-batch $\mathcal{B}$ is
% \vspace*{-0.05in}
\begin{equation}
    \begin{aligned}
     &\nabla_{\theta_s}\widehat{\Phi}(\theta_s^t)=- \sum_{i=1}^\mathcal{B} P_i \nabla_{\theta_s} h(\theta_s^t; \xi_i)\nabla_{\theta_p}\mathcal{L}(\theta_p^t; \xi_i)^Tz^t \\
     &+\sum_{i=1}^\mathcal{B} P_i \sum_{j=1}^\mathcal{B}  P_j \nabla_{\theta_s} h(\theta_s^t; \xi_j)\nabla_{\theta_p} \mathcal{L}(\theta_p^t; \xi_i)^Tz^t.
    \end{aligned} \label{ditributed_softmax}
    % \vspace*{-0.08in}
\end{equation}
% \vspace*{-0.05in}
Then the update for the upper-level variable ($\theta_s$) is  
$\theta_s^{t+1} = \theta_{s}^t - \alpha \nabla_{\theta_s}\widehat{\Phi}(\theta_s^t).$
% \vspace*{-0.05in}
When the score model converges over $T$ steps, reaching $\theta_s^T$, it is then used to estimate the influence scores of the entire training dataset in the current round by: $ S_i = h(\theta_s^T, \xi_i), \, \forall \xi_i \in \mathcal{D}_{tr}.$
Then the influence scores are collected: $\{S_i\mid 0\leq i\leq |D_{tr}| \}$, and the top-ranked samples with the highest influence scores are selected to construct $\mathcal{D}_s$, which is used to pretraining the LLM ($\theta_{tr}$).

The detailed implementation of the algorithm is presented in  Algorithm \ref{alg:alg1}. In practice, we call SGD optimizer \texttt{SGD(variable, gradient, lr, steps)} to update the lower-level variable for $N$ steps with learning rate \texttt{lr}. We use Adam optimizer (\texttt{Adam(variable, gradient, lr)}) \citep{kingma2014adam} to update the upper-level variables.
The pretraining process is conducted over $R$ rounds. In each round, the algorithm performs data selection followed by LLM retraining. The training dataset is partitioned into $R$ shards. The data selection in round $r$ is conducted on $\mathcal{D}_{tr}^r$. The LLM resumes training from the previous round’s checkpoint and updates to $\theta_{tr}^{r}$ at the end of the $r$-th round. Similarly, the score model also continues learning throughout the process, reaching  $\theta_{s}^r$ at the $r$-th round. 
% It is worth noting that the proxy model ($\theta_{p}^r$) is reinitialized with the warm-up model at the beginning of each round. This prevents the model from overfitting to the previous round’s training data and ensures it can better capture the evolving behavior of the LLM. 

It is worth noting that the proxy model $\theta_p$ is reinitialized with the
warm-up proxy at the beginning of each round, while the target LLM continues
training from the previous round. 
The data distribution changes substantially from one round to the next (i.e., $\xi$ comes from a different data distribution when the round changes), which leads to a completely different lower-level problem. As a result, simply warm-starting as \citep{ji2021bilevel} from the proxy of the previous round may introduce a strong bias toward the previous selected distribution. In our setting, this can reduce the proxy's ability to adapt to the current round, i.e., its \emph{plasticity}. This intuition is consistent with prior work \citep{shin2024dash} on loss of plasticity, which shows that under non-stationary training distributions, a network trained for a long time can become harder to adapt than a freshly initialized counterpart. Therefore, resetting the proxy $\theta_p$ with random initialization at each round is intended to let it re-fit the newly data distribution with high plasticity and keep alignment with the target LLM, rather than inheriting optimization bias from earlier rounds.

% Since each round operates on a different data shard and the selected-data distribution may change across rounds, reusing the
% previous proxy can bias the lower-level optimization toward outdated data
% preferences. Resetting the proxy helps preserve its plasticity for the current
% round, while the cross-round evolution of the training state is carried by the
% continuously updated target LLM and score model.

% The details about the warm-up model is presented in Section~\ref{sec:warm-up}.

\begin{algorithm}[!t]
    \caption{\texttt{BLISS} }\label{alg:alg1} %: Bilevel Influence Scoring Method for Data Selection
    \begin{algorithmic}[1]
        \STATE \textbf{Input:} $ \eta_1, \eta_2, \eta_3, \alpha,  R, T, K, Q, N, \mathcal{D}_{tr},  \tilde{\mathcal{D}}_{tr}, \mathcal{D}_{val}$ 
        \STATE \textbf{Initialize:} Warm up $\theta_p^{0, 0}, \theta_{s}^{0, 0}, \theta_{tr}^{0, 0}$ using randomly selected training data.
        \FOR {$r = 0, \dots, R-1$ }  
            \STATE \# \textit{reset proxy/score parameters for a new round} 
            \STATE $\theta_p^{0, r} = \theta_p^{0, 0}$ 
            \STATE $\theta_s^{0, r} = \theta_s^{T, r-1}$ if $r \ge  1$ else  $\theta_s^{0, 0}$
            \STATE $\theta_{tr}^{0, r} = \theta_{tr}^{Q, r-1}$ if $r\ge1$ else $\theta_{tr}^{0, 0}$ 
        \FOR{$t=0, \dots, T-1$} 
            \STATE Sample $\xi_t^r, \tilde{\xi}_t^r, \pi_t^r \leftarrow \tilde{\mathcal{D}}_{tr}^r$, and sample $\zeta_t \leftarrow \mathcal{D}_{val}$
            \STATE \#  \textit{LL: update the proxy model for $N$ steps}
            \STATE  $\theta_p^{t+1, r} = \texttt{SGD}(\theta_p^{t, r}, \nabla_{ \theta_p} G(\theta_p^{t, r}, \theta_s^{t, r}; \xi_t^r), \eta_1, N)$
            \STATE \#  \textit{solve the linear system}
            \STATE  $\begin{aligned}[t]
                z^{t+1, r} = \texttt{GDLS}(\eta_2, K, &\nabla_{\theta_p} G(\theta_p^{t, r}, \theta_{s}^{t, r}; \tilde{\xi}_t^r), \\
                &\nabla_{\theta_p}F(\theta_p^{t, r}, \theta_{s}^{t, r}; \zeta_t)) 
                \end{aligned}$   
            \STATE            \#  \textit{UL: update the score model }
            \STATE  $\begin{aligned}
            \theta_{s}^{t+1, r} = \texttt{Adam} (\theta_{s}^{t, r}, -\nabla_{\theta_s\theta_p}^2G&(\theta_p^{t+1, r}, \theta_s^{t, r}; \\ & \pi_t^r )z^{t+1, r}, \alpha)\end{aligned}$ \footnotemark\hfill
 
        \ENDFOR
        \STATE Infer the influence score $\{S_i^r\mid 0 \leq i \leq |\mathcal{D}_{tr}^r|-1 \}$ on $\mathcal{D}_{tr}^r$ using $\theta_s^{T, r}$
        \STATE Sort $\{S_i^r\}$ in descending order and select the $20\%$ data with the highest influence scores from $\mathcal{D}_{tr}^r$ to form the selected data $\mathcal{D}_{s}$
         \FOR{$\tau=0, \dots, Q-1$} 
            \STATE  Sample $\xi_{\tau}$ from $\mathcal{D}_s$.
            \STATE \# \textit{pretrain the LLM}
            \STATE $\theta_{tr}^{\tau+1, r} =\theta_{tr}^{\tau, r} - \eta_3  \nabla_{ \theta_{tr}} \mathcal{L}( \theta_{tr}^{\tau, r}; \xi_{\tau})$  
         \ENDFOR
        \ENDFOR
    \end{algorithmic}
\end{algorithm}
\vspace{-0.05in}   
\begin{algorithm}[!t]
    \caption{\texttt{GDLS: Gradient Descent for the Linear System Solution}} \label{alg:alg2}
    \begin{algorithmic}[1]
        \STATE \textbf{Input:} $ \eta, K, \nabla_{\theta_p}g(\theta_p), a$
         \STATE \textbf{Initialize:} $z_0$
        \FOR{$k=0, \dots, K-1$} 
        % \STATE $z_k =z_{k-1}$ if $k\ge1$ else $z_0$ 
        \STATE $z_{k+1} = z_k - \eta \big( \nabla_{\theta_p}^2g(\theta_p) z_k - a\big)$
        \ENDFOR
    \STATE Return $z_K$

    \end{algorithmic}
\vspace{-0.05in}    
\end{algorithm}
% \vspace*{-0.05in}

\subsection{Warm Up Models} \label{sec:warm-up}
% \vspace*{-0.05in}

The key distinction between our algorithm and other data selection methods \citep{brown2020language,xie2023data,wettig2024qurating} is that it operates independently of external pretrained models, avoiding biases from data selection influenced by such models. However, without leveraging pretrained knowledge, the proxy model, score model, and LLM tend to perform poorly in the initial phase due to random parameter initialization. To mitigate this issue, we incorporate a model warm-up step before data selection, similar to other data selection approaches \citep{yu2024mates, xia2024less}, using randomly selected samples. The lightweight proxy and score models share token embedding layers and transformer blocks but differ in their final layers: the proxy model handles token generation, while the score model outputs influence scores for individual samples. Consequently, only the proxy model and the LLM require warm-up, while the score model can be initialized with the weights from proxy model directly.

We provide an additional convergence analysis of our algorithm in
\cref{app:analysis}, which clarifies the role of the proxy update
steps, the selection ratio, and the round-wise proxy reset.

\section{Experiments}\label{sec:exp}
% \footnotetext{\texttt{Adam(variable, gradient, lr)} optimizer receives the current variable, its hypergradient and learning rate. Then it updates the first and second momentum, then returns the updated variable.}
In this section, we validate the proposed bilevel influence scoring framework for pretraining data selection. We apply the bilevel optimization algorithm to train a lightweight proxy model ($\theta_p$) and a score model ($\theta_s$)) for data selection. We then pretrain a target LLM ($\theta_{tr}$), specifically Pythia-410M/1B, from scratch on a selected subset of the large-scale C4 dataset \citep{raffel2020exploring}, which is designed for LLM pretraining.
we then evaluate the pretrained LLM on multiple downstream tasks and compare its performance against several baseline methods, including Random selection, DSIR \citep{xie2023data}, SemDeDup \citep{abbas2023semdedup}, DsDm \citep{engstrom2024dsdm}, LESS \citep{xia2024less}, QuRating \citep{wettig2024qurating}, and MATES \citep{yu2024mates}. We furthermore scale up our experiment to 2.8B model pretraining and achieve $1.4\%$ performance improvement over the state-of-the-art method. We also verify the the fidelity of proxy models to full-scale LLMs by domain reweighting experiment on SlimPajama-6B, deferred to \cref{sec:domain_reweight}. The code is available at \url{https://github.com/MingruiLiu-ML-Lab/BLISS-Bilevel-Data-Selection}.

\subsection{Dataset Settings}

Following the approach of DsDm \citep{engstrom2024dsdm}, we perform data selection and pretraining using tokenized data. The procedure of BLISS is implemented for 5 rounds (i.e., $R=5$),with the C4 dataset partitioned into five equal shards, denoted as $\{\mathcal{D}_{tr}^r \mid 0\leq r \leq4\}$. Each training round operates on a distinct data shard without replacement. In every round, we first uniformly sample a small proportion ($0.1\%$) from $\mathcal{D}_{tr}^r$ as the bilevel training set $\tilde{\mathcal{D}}_{tr}^r$ for updating the proxy model. We choose LAMBADA \citep{paperno2016lambada} as validation data for updating the score model as the prior work \citep{engstrom2024dsdm,yu2024mates}. LAMBADA requires broad discourse
context for word prediction and therefore provides a semantically demanding
upper-level signal beyond local next-token prediction. More details about the choice of validation set can be found in \cref{sec:val_data}.
Other datasets, including ARC-E \citep{clark2018think}, SQUAD \citep{rajpurkar2016squad}, and PIQA \citep{bisk2020piqa}, are  evaluated in the ablation study (\Cref{sec:val_data}). 

To evaluate the performance of data selection algorithms, we run the pretraining model across 9 downstream tasks, including SciQ \citep{welbl2017crowdsourcing}, ARC-E \citep{clark2018think}, ARC-C \citep{clark2018think}, LogiaQA \citep{liu2020logiqa}, OBQA \citep{mihaylov2018can}, BoolQ \citep{clark2019boolq}, HellaSwag \citep{zellers2019hellaswag}, PIQA \citep{bisk2020piqa}, and WinoGrande \citep{sakaguchi2021winogrande}. These tasks cover a diverse range of reasoning and comprehension challenges, including question answering, logical inference, commonsense reasoning, and coreference resolution. Thus it requires models to demonstrate various capabilities, such as retrieving and applying scientific knowledge, understanding causal relationships, resolving ambiguities in natural language, and making informed choices among distractors. A good data selection algorithm is expected to select the "important" data that boost model performance across these downstream tasks.

\subsection{Model Settings}

The target pretraining model, Pythia-410M/1B/2.8B, consists of 410 million, 1 billion or 2.5 billion trainable parameters. Both the proxy model and score model are based on Pythia-31M (for Pythia-410M) or Pythia-160M (for Pythia-1B), but they serve different purposes: the proxy model acts as a surrogate for the LLM and is trained for next-token prediction, while the score model functions as a regression model that maps individual samples to corresponding influence scores. Details of model settings are deferred to Appendix \ref{sec:model_setting}. Notably, all models are trained from scratch using Gaussian initialization for model parameters. Additional experimental details, including hyperparameter choices, learning rate schedules, and distributed training strategies, are provided in Appendix \ref{sec:exp_setting}.

\subsection{Bilevel Optimization for Proxy Model and Score Model}

In the Pythia-410M setting, the proxy model $\theta_p$
is updated with a ``single-step" ($N=1$) optimization per iteration (line 9 in Algorithm \ref{alg:alg1}). However, when scaling up to larger models like Pythia-1B, we adopt a ``multi-steps" ($N=5$)  update strategy for the proxy model to achieve a better lower-level solution.
To demonstrate the effectiveness of bilevel optimization in training the proxy model and score model, we visulize the evolution of the training loss at during round 2 (Figure \ref{fig:bilevel_training_r2} in Appendix \ref{loss_fig}) and round 5 (Figure \ref{fig:bilevel_training_r5} in Appendix \ref{loss_fig}). Since the first round uses randomly selected data to warm up the LLM, our data selection algorithm is employed from the second round onward.

 Within each round, both losses exhibit a two-phase trend: they initially decrease rapidly before experiencing a slight increase. This behavior arises due to the composition of the lower-level objective function, which includes three terms: the weighted cross-entropy loss, the KL divergence loss, and a regularization term. In the first phase, the weighted cross-entropy loss dominates, decreasing as the proxy model is optimized. In the second phase, the KL divergence term becomes more influential. Since the LLM has not yet been trained on the current dataset $\mathcal{D}_{tr}^{r}$ (it only performs inference in bilevel training), its predictions may be suboptimal. The KL divergence term encourages the proxy model to mimic the behavior of this "imperfect" LLM, leading to a slight performance degradation. However, this ensures that the proxy model's data preference aligns with that of the LLM, improving the relevance of the selected training data and ultimately boosting the LLM’s downstream task performance. An ablation study on the effect of KL divergence loss is presented in Section \ref{sec:kl_div}.

From round 2 to round 5, the score model is continuously optimized, leading to more accurate sample weight assignments. This, in turn, enhances the proxy model’s performance on the weighted training samples, further improving the quality of data selection.

\begin{table*}[!t]
    \centering
    % \vspace{-0.1in}
      \setlength{\tabcolsep}{1pt}
    
    \caption{Comparison of methods on zero-shot evaluation over multiple downstream datasets (410M/1B model, 25B tokens data). Best results are marked bold. The accuracy with standard error is reported based on the lm-evaluation-harness \citep{gao2021framework} implementation.}
    % \vspace{-0.05in}
    \resizebox{0.95\textwidth}{!}{
    \begin{tabular}{lcccccccccc}
        \toprule
        \textbf{Methods} (\#FLOPs $\times 10^{19}$) & \textbf{SciQ} & \textbf{ARC-E} & \textbf{ARC-C} & \textbf{LogiQA} & \textbf{OBQA} & \textbf{BoolQ} & \textbf{HellaSwag} & \textbf{PIQA} & \textbf{WinoGrande} & \textbf{Average} \\
        \midrule
        \multicolumn{11}{l}{\textbf{410M Setting:} 410M model, 25B tokens} \\
        \midrule
         Random  (6.35) & 64.1 \scriptsize{(1.5)} & 40.2 \scriptsize{(1.0)} & \textbf{25.6} \scriptsize{(1.3)} & 24.7 \scriptsize{(1.7)} & 29.4 \scriptsize{(2.0)} & 58.9 \scriptsize{(0.9)} & 39.7 \scriptsize{(0.5)} & 67.1 \scriptsize{(1.1)} & 50.6 \scriptsize{(1.4)} & 44.5 \scriptsize{(1.3)} \\
        MATES  (8.11) & 65.7 \scriptsize{(1.5)} & 41.5 \scriptsize{(1.0)} & 25.0 \scriptsize{(1.3)} & 26.1 \scriptsize{(1.7)} & \textbf{30.8} \scriptsize{(2.1)} & \textbf{60.6} \scriptsize{(0.9)} & 41.0 \scriptsize{(0.5)} & 67.8 \scriptsize{(1.1)} & 51.8 \scriptsize{(1.4)} & 45.7 \scriptsize{(1.4)} \\
        BLISS (8.08)   & \textbf{68.1} \scriptsize{(1.5)} & \textbf{42.2} \scriptsize{(1.0)} & 25.1 \scriptsize{(1.3)} & \textbf{27.3} \scriptsize{(1.7)} & 29.6 \scriptsize{(2.0)} & 59.3 \scriptsize{(0.9)} & \textbf{41.2} \scriptsize{(0.5)} & \textbf{68.2} \scriptsize{(1.1)} & \textbf{52.0} \scriptsize{(1.4)} & \textbf{45.9} \scriptsize{(1.4)}\\
        \bottomrule
        \multicolumn{11}{l}{\textbf{1B Setting:} 1B model, 25B tokens} \\
        \midrule
        Random {(17.67)} & 65.8\scriptsize{(1.5)} & 43.7\scriptsize{(1.0)} & 25.6\scriptsize{(1.3)} & 27.5\scriptsize{(1.8)} & 31.8\scriptsize{(2.1)} & 60.2\scriptsize{(0.9)} & 43.8\scriptsize{(0.5)} & 68.9\scriptsize{(1.1)} & 50.7\scriptsize{(1.4)} & 46.4\scriptsize{(1.4)} \\
        MATES {(19.97)} & 67.3\scriptsize{(1.5)} & 44.9\scriptsize{(1.0)} & \textbf{25.9}\scriptsize{(1.3)} & \textbf{28.7}\scriptsize{(1.8)} & 32.2\scriptsize{(2.1)} & \textbf{60.9}\scriptsize{(0.9)} & 45.3\scriptsize{(0.5)} & 69.5\scriptsize{(1.1)} & 52.4\scriptsize{(1.4)} & 47.5\scriptsize{(1.4)} \\
        BLISS {(19.53)}   & \textbf{69.4}\scriptsize{(1.5)} & \textbf{45.7}\scriptsize{(1.0)} & 24.8\scriptsize{(1.3)} & {25.8}\scriptsize{(1.7)} & \textbf{33.2}\scriptsize{(2.1)} & 59.8\scriptsize{(0.9)} & \textbf{47.8}\scriptsize{(0.5)} & \textbf{71.6}\scriptsize{(1.1)} & \textbf{52.9}\scriptsize{(1.4)} & \textbf{47.9}\scriptsize{(1.3)}\\
        \bottomrule
    \end{tabular}}\label{tbl:bliss_vs_mates}
\end{table*}

\subsection{Evaluation Results on the Downstream Tasks}
\begin{figure}[t]
    \centering
    \begin{subfigure}{\linewidth}
        \centering
        \includegraphics[width=0.47\linewidth]{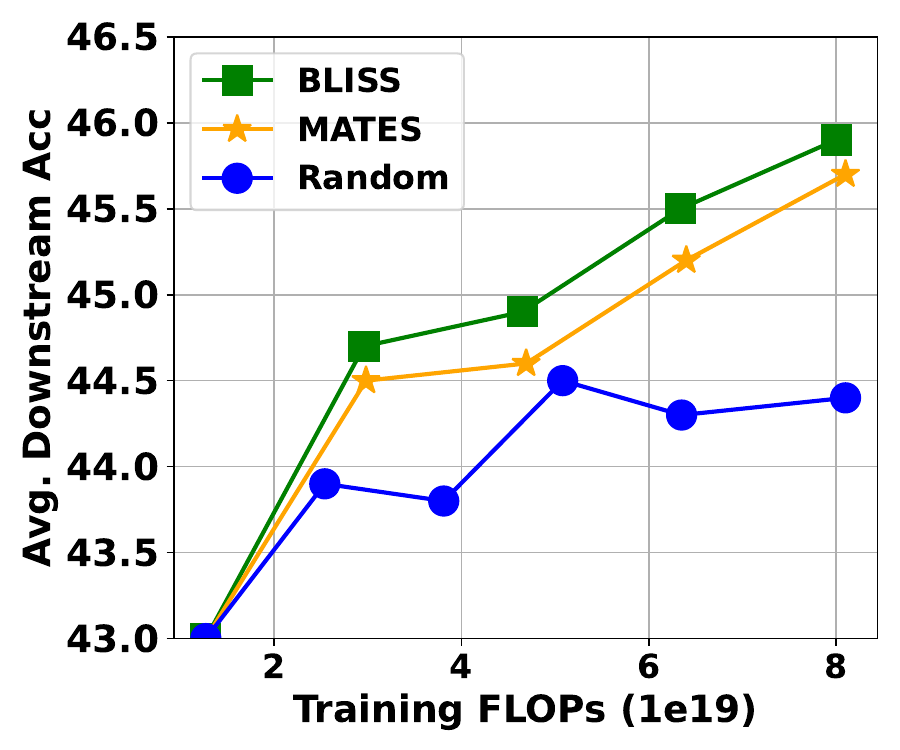}
        \includegraphics[width=0.47\linewidth]{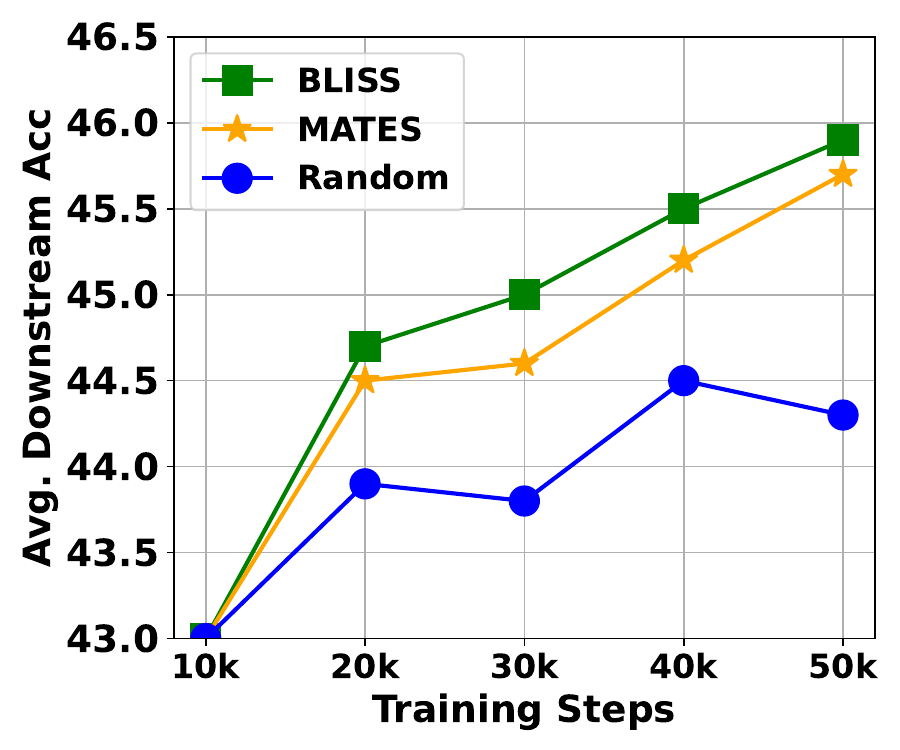}
        \caption{Pretraining the 410M model.}
        \label{fig:acc_410m}
    \end{subfigure}
    \hfill
    \begin{subfigure}{\linewidth}
        \centering
        \includegraphics[width=0.47\linewidth]{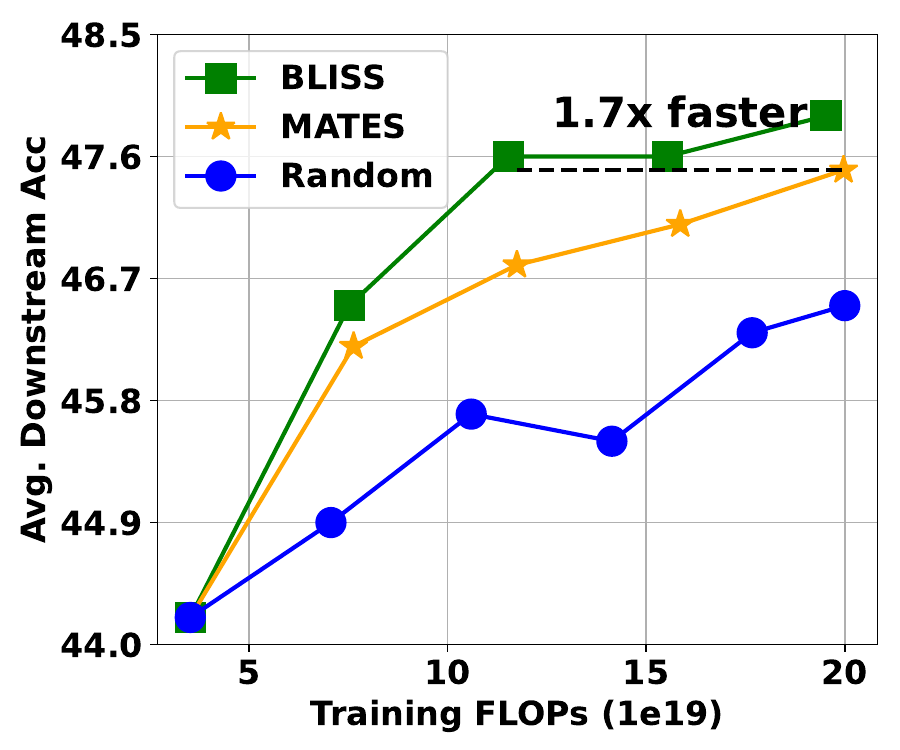}
        \includegraphics[width=0.47\linewidth]{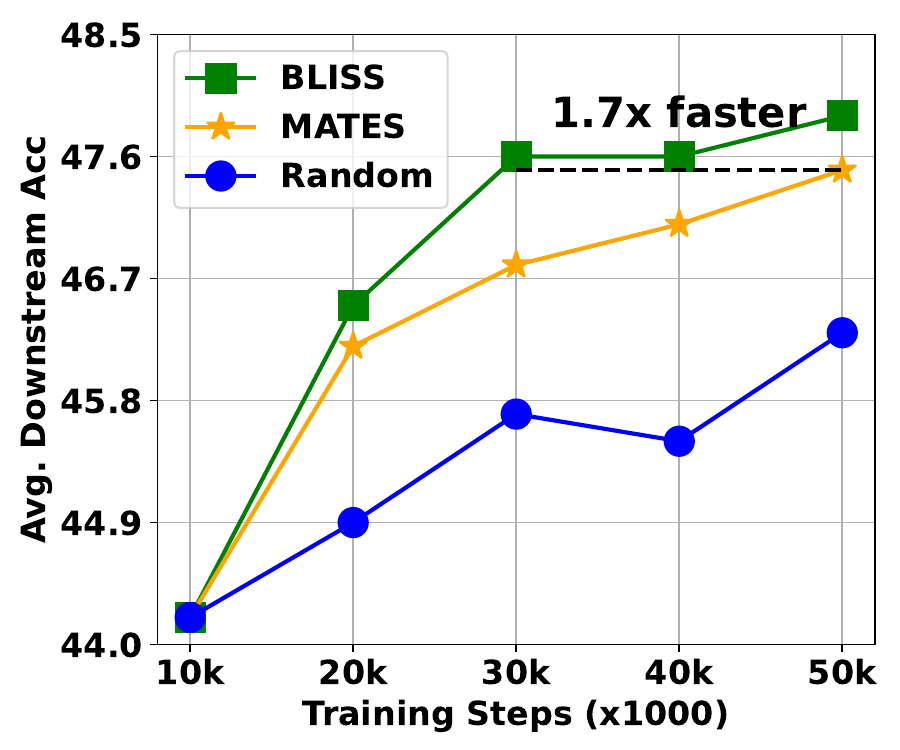}
        \caption{Pretraining the 1B model.}
        \label{fig:acc_1b}
    \end{subfigure}
    \caption{Downstream performance of Pythia-410M and Pythia-1B with respect to pretraining FLOPs and steps. The first point corresponds to a warm-up model trained on random data.}
    \label{fig:acc_step}
    \vspace{-0.2in}
\end{figure}

The LLM is continuously trained for 10,000 steps on the selected data in each round. After completing five rounds of training, we evaluate the zero-shot performance of Pythia-410M/1B on various downstream tasks and report the average accuracy along with the standard error for each dataset(see Table \ref{tbl:bliss_vs_mates}. Our algorithm consistently outperforms MATES and random selection methods across multiple tasks.  For example on 410M setting, BLISS, compared with MATES, improves $2.4\%$ on SciQ, $0.7\%$ on ARC-E, $0.8\%$ on LogiQA, $0.2\%$ on HellaSwag, $0.4\%$ on PIQA, $0.2\%$ on WinoGrande, and $0.2\%$ on average accuracy (see Table~\ref{tab:mates_results}). Additionally, Figure \ref{fig:acc_step} presents the evaluation results in relation to pretraining FLOPs and training steps. BLISS consistently outperforms other baseline methods throughout the entire five-round pretraining process (with 10k steps per round). In particular, our method on 1B setting achieves a $1.7\times$ speedup in reaching the same performance as MATES, further validating the effectiveness of our data selection approach.

% \begin{table}[!t]
% % \vspace{-0.1em}
% \centering
% \setlength{\tabcolsep}{4pt}
% % \vspace*{-0.05in}

% \begin{minipage}[t]{0.48\textwidth}
% \centering
% \captionof{table}{Average evaluation accuracy (15B tokens data) by pretraining 2.8B model with data selected from the 1B model experiment.}
% \label{tbl:consistency}
% % \vspace{-0.1in}
% \resizebox{1\linewidth}{!}{
% \setlength{\tabcolsep}{15pt}
% \begin{tabular}{lccc}
% \toprule
% \textbf{Methods} & \textbf{Round 1 (Random)} & \textbf{Round 2} & \textbf{Round 3} \\
% \midrule
% MATES & 45.9 {\scriptsize(1.3)} & 47.4 {\scriptsize(1.3)} & 47.6 {\scriptsize(1.3)} \\
% BLISS & 45.2 {\scriptsize(1.3)} & \textbf{47.6} {\scriptsize(1.3)} & \textbf{49.0} {\scriptsize(1.3)} \\
% \bottomrule
% \end{tabular}}
% \end{minipage}\hfill
% \begin{minipage}[t]{0.48\textwidth}
% \centering
% \captionof{table}{Average evaluation accuracy of 3 rounds by pretraining Llama-0.5B model. Llama-134M is deployed as the proxy model in BLISS.}
% \label{tbl:llama}

% \resizebox{1\linewidth}{!}{
% \setlength{\tabcolsep}{15pt}
% \begin{tabular}{lccc}
% \toprule
% \textbf{Methods} & \textbf{Round 1 (Random)} & \textbf{Round 2} & \textbf{Round 3} \\
% \midrule
% MATES & 43.12 {\scriptsize(1.27)} & 44.53 {\scriptsize(1.27)} & 45.01 {\scriptsize(1.27)} \\
% BLISS & 43.12 {\scriptsize(1.27)} & \textbf{44.57} {\scriptsize(1.27)} & \textbf{45.65} {\scriptsize(1.27)} \\
% \bottomrule
% \end{tabular}}
% \end{minipage}
% % \vspace{-0.05in}
% \end{table}

\begin{table}[t]
\centering
\caption{Average evaluation accuracy (15B tokens) by pretraining the 2.8B model with data selected from the 1B-model experiment.}
\label{tbl:consistency}
\resizebox{\linewidth}{!}{
\begin{tabular}{lccc}
\toprule
\textbf{Methods} & \textbf{Round 1 (Random)} & \textbf{Round 2} & \textbf{Round 3} \\
\midrule
MATES & 45.9 {\scriptsize(1.3)} & 47.4 {\scriptsize(1.3)} & 47.6 {\scriptsize(1.3)} \\
BLISS & 45.2 {\scriptsize(1.3)} & \textbf{47.6} {\scriptsize(1.3)} & \textbf{49.0} {\scriptsize(1.3)} \\
\bottomrule
\end{tabular}}
\vspace{-0.05in} 
\end{table}

\begin{table}[t]
\centering
\caption{Average evaluation accuracy over three rounds by pretraining the LLaMA-0.5B model. LLaMA-134M is used as the proxy model in BLISS.}
\label{tbl:llama}
\resizebox{\linewidth}{!}{
\begin{tabular}{lccc}
\toprule
\textbf{Methods} & \textbf{Round 1 (Random)} & \textbf{Round 2} & \textbf{Round 3} \\
\midrule
MATES & 43.12 {\scriptsize(1.27)} & 44.53 {\scriptsize(1.27)} & 45.01 {\scriptsize(1.27)} \\
BLISS & 43.12 {\scriptsize(1.27)} & \textbf{44.57} {\scriptsize(1.27)} & \textbf{45.65} {\scriptsize(1.27)} \\
\bottomrule
\end{tabular}}
\vspace{-0.05in} 
\end{table}

\paragraph{Scaling Up to 2.8B Model Pretraining using the Data Selected by 160M/1B Experiment.} To further validate the selected data is of good quality regardless of model size, we pretrain a larger model of 2.8B parameters with data selected from the 1B model experiment with 160M proxy and score models. We run MATES and BLISS for 3 rounds (15B tokens). As shown in Table~\ref{tbl:consistency}, BLISS consistently outperforms MATES across all data selection rounds, achieving $1.4\%$ accuracy improvement over MATES in round 3.

\paragraph{Generalize model architecture to LLaMA family.} We also explore LLaMA architecture models to validate the generalization of our method. Specifically, we use LLaMA-0.5B as the target pretraining model, and LLaMA-134M as the proxy model and score model. In each round, we first minimize the difference between the proxy model and the target model by training the proxy model toward a lower KL divergence. Then we periodically reset the proxy model to the initial state, in addition to resetting it at the beginning of each round. Table \ref{tbl:llama} presents the evaluation results compared with MATES, where BLISS exhibits strong data selection performance. At the round $3$, our algorithm improves over MATES by $0.6\%$. 
More details are presented in Appendix \ref{sec:llama}. 

\subsection{Computational Cost} \label{sec:cost}

We follows the FLOPs estimation method in \citet{li2024datacomp} and report the total GPU FLOPs, including the pretraining, model warm-up, and data selection.  Our main observation is: without relying on any external pretrained models as required in MATES, BLISS achieves higher average downstream performance while consuming fewer FLOPs. A detailed comparison of total FLOPs consumption is provided in Table \ref{tab:flops_comparison}, and running time/memory comparison is presented in \cref{sec:runtime_memory}.

With the same pretraining budget for LLM and an equivalent number of training tokens, BLISS is more efficient in data selection than MATES. The higher computational cost of MATES is due to its reliance on oracle data influence estimation, which involves computing the loss change after performing a one-step gradient descent update on an individual training sample. This process is highly time-consuming, because it requires per-sample gradient and cannot increase the batch size per GPU.
In contrast, BLISS formulates data selection as a bilevel optimization problem, enabling the lightweight score model and proxy model to be trained to convergence within relatively few steps, i.e., 3,000 per round (ablation study for bilevel steps is presented in \cref{sec:bilevel_step}). While BLISS introduces additional training steps for warming up the proxy and score models from scratch, this cost is negligible compared to the overall pretraining FLOPs. 
One limitation of BLISS is that the HVP computation in bilevel optimization
incurs higher peak memory usage than MATES, as reported in \cref{tbl:time_mem}. This
represents a memory-time trade-off: BLISS requires more peak memory, but its
data-selection stage is substantially faster because it avoids per-sample
one-step influence estimation on the target LLM.

% One limitation of BLISS we have to acknowledge is that HVP calculation in bilevel optimization incurs higher peak memory usage, see \cref{tbl:time_mem} for details.   

\begin{table*}[!t]
    \centering
    % \vspace{-0.02in}
    \renewcommand{\arraystretch}{1.2}  % Increase row height
        \caption{Total FLOPs for pretraining 410M/1B model with 25B tokens.}
        % \vspace{-0.1in}
    \resizebox{0.9\textwidth}{!}{
     \renewcommand{\arraystretch}{1.0}
    \begin{tabular}{lcc|lcc}
        % \toprule
        \Xhline{1.2pt}
        \textbf{Model} & \textbf{\#FLOPs $\times 10^{19}$} & \textbf{Ratio} &\textbf{Model} & \textbf{\#FLOPs $\times 10^{19}$} & \textbf{Ratio} \\
        \hline
        \multicolumn{3}{l|}{\textbf{MATES:} 410M model, 25B tokens}  & \multicolumn{3}{r}{\textbf{BLISS:} 410M model, 25B tokens}\\
        \hline
        Model pretraining             & 6.35  & 78.3\%          & Model pretraining  & 6.35  & 78.59\%\\
        Oracle data influence collection & 0.29  & 3.58\%      & Warm up the proxy/score model  & 0.07 & 0.87\% \\
        Data influence model training  & 0.01  & 0.1\%          & Bilevel optimization & 0.13  & 1.62\%\\
        Data influence model inference & 1.46  & 18.0\%         & Data influence model inference & 1.53  & 18.94\% \\
        \textbf{Total}     & 8.11  & 100.00\%                   & \textbf{Total}         & \textbf{8.08}  & 100.00\% \\
        \hline
        \multicolumn{3}{l|}{\textbf{MATES:} 1B model, 25B tokens}  & \multicolumn{3}{r}{\textbf{BLISS:} 1B model, 25B tokens} \\
        \hline
        Model pretraining             & 17.67  & 88.5\%         & Model pretraining   & 17.67  & 90.48\%\\
        Oracle data influence collection  & 0.83 & 4.1\%        & Warm up the proxy/score model & 0.07   & 0.36\%\\
        Data influence model training          & 0.01  & 0.1\%  & Bilevel optimization   & 0.261  & 1.34\%\\
        Data influence model inference & 1.46  & 7.3\%          & Data influence model inference & 1.53  & 7.83\% \\
        \textbf{Total} & 19.97  & 100.00\%    & \textbf{Total}  & \textbf{19.53}   & 100.00\% \\
        \Xhline{1.2pt}
    \end{tabular}}
    \label{tab:flops_comparison}
    % \vspace{-0.1in}
\end{table*}

% \begin{figure}[!t]
%     \centering
%     \subfigure[Pretrain 410M model]{
%     \includegraphics[width=0.23\linewidth]{figures/acc_flops_410m.pdf}
%     \includegraphics[width=0.23\linewidth]{figures/acc_steps_410m.pdf}}
%     \subfigure[Pretrain 1B model]{
%     \includegraphics[width=0.23\linewidth]{figures/acc_flops_1b.pdf}
%     \includegraphics[width=0.23\linewidth]{figures/acc_steps_1b.pdf}}
%     \vspace{-0.1in}
%     \caption{The downstream performance of Pythia-410M/1B model w.r.t. pretraining FLOPs and steps, where the first point denotes the performance of a warm-up model trained on random data.}
%     \label{fig:acc_step}
% \end{figure}

\section{Ablation Studies}

To inspect the effectiveness of key techniques used in our proposed algorithm, we conduct ablation studies on the effect of bilevel optimization (Section \ref{sec:bilevel_opt}), KL divergence loss (Section \ref{sec:kl_div}), the impact of softmax reparameterization on the score model’s outputs (Appendix \ref{sec:softmax_score}), the size of proxy model (Appendix \ref{sec:model_size}), the initialization for the score model (Appendix \ref{sec:init_method}), and the influence of different validation datasets ($\mathcal{D}_{val}$) on performance (Appendix \ref{sec:val_data}).

% \vspace{-0.2in}
\subsection{Single-level versus Bilevel Optimization} \label{sec:bilevel_opt}

In bilevel algorithm, the hyper-gradient is essential for the update of upper level parameters. To verify the effectiveness of bilevel update for the upper-level parameters, we compare bilevel update with a single update, which update $\theta_s$ and $\theta_p$ together using both training and validation data for the lower-level objective. Specifically, the upper and lower levels are reduced to a single level problem: the upper-level and lower-level parameters are updated simultaneously on validation dataset and training dataset respectively. With the same number of training steps as bilevel training, the average accuracy of single level update degrades $0.5\%$ as shown in Table \ref{tbl:single_level} in \cref{sec:mode_ablation}.

\subsection{ Alignment via KL Divergence} \label{sec:kl_div}

Our objective is to select training data that maximizes the LLM's performance on downstream tasks. To achieve this, the proxy model must effectively represent the LLM, which we enforce by applying KL divergence loss to align their output logits. 
As shown in Figure \ref{fig:training_without_kl2} (Appendix \ref{sec:mode_ablation}), incorporating KL divergence leads to improved performance across most downstream tasks, with a 9.3\% accuracy boost on LogiQA and a 1.4\% increase in average accuracy. Interestingly, while removing KL divergence results in a lower training loss (as seen in Figure \ref{fig:training_without_kl1} compared to Figure \ref{fig:bilevel_training_r2} in Appendix \ref{loss_fig}), it does not translate to better downstream performance. These findings highlight the importance of bridging the gap between the proxy model and the LLM to ensure effective data selection, demonstrating that a closer alignment between the two models leads to better overall performance. 

\section{Conclusion}

In this paper, we present BLISS, a lightweight bilevel influence scoring method for data selection in language model pretraining. BLISS utilizes a proxy model, a score model, and a novel bilevel optimization framework to model dynamic data preference under model updates without relying on external pretrained models. Experimental results demonstrate its effectiveness in selecting data for pretraining Pythia and LLaMA models. However, the bilevel optimization in our method may incur higher peak memory usage, which remains an important direction for future work.

\section*{Acknowledgements}
This work is supported by NSF award \#2436217,
\#2425687, \#2613303, \#2228603, \#2411248.
We gratefully acknowledge the computational resources provided by the IBM T.J. Watson Research Center. We also acknowledge the computational support from the NSF NAIRR project under NAIRR award \#250070.

\section*{Impact Statement}
This paper presents work whose goal is to advance the field of machine learning. There are many potential societal consequences of our work, none of which we feel must be specifically highlighted here.

\bibliography{ref}

@article{paperno2016lambada,
  title={The LAMBADA dataset: Word prediction requiring a broad discourse context},
  author={Paperno, Denis and Kruszewski, Germ{\'a}n and Lazaridou, Angeliki and Pham, Quan Ngoc and Bernardi, Raffaella and Pezzelle, Sandro and Baroni, Marco and Boleda, Gemma and Fern{\'a}ndez, Raquel},
  journal={arXiv preprint arXiv:1606.06031},
  year={2016}
}

@article{shen2024seal,
  title={Seal: Safety-enhanced aligned llm fine-tuning via bilevel data selection},
  author={Shen, Han and Chen, Pin-Yu and Das, Payel and Chen, Tianyi},
  journal={arXiv preprint arXiv:2410.07471},
  year={2024}
}

@article{raffel2020exploring,
  title={Exploring the limits of transfer learning with a unified text-to-text transformer},
  author={Raffel, Colin and Shazeer, Noam and Roberts, Adam and Lee, Katherine and Narang, Sharan and Matena, Michael and Zhou, Yanqi and Li, Wei and Liu, Peter J},
  journal={Journal of machine learning research},
  volume={21},
  number={140},
  pages={1--67},
  year={2020}
}

@article{chowdhery2023palm,
  title={Palm: Scaling language modeling with pathways},
  author={Chowdhery, Aakanksha and Narang, Sharan and Devlin, Jacob and Bosma, Maarten and Mishra, Gaurav and Roberts, Adam and Barham, Paul and Chung, Hyung Won and Sutton, Charles and Gehrmann, Sebastian and others},
  journal={Journal of Machine Learning Research},
  volume={24},
  number={240},
  pages={1--113},
  year={2023}
}

@article{elazar2023s,
  title={What's In My Big Data?},
  author={Elazar, Yanai and Bhagia, Akshita and Magnusson, Ian and Ravichander, Abhilasha and Schwenk, Dustin and Suhr, Alane and Walsh, Pete and Groeneveld, Dirk and Soldaini, Luca and Singh, Sameer and others},
  journal={arXiv preprint arXiv:2310.20707},
  year={2023}
}

@inproceedings{du2022glam,
  title={Glam: Efficient scaling of language models with mixture-of-experts},
  author={Du, Nan and Huang, Yanping and Dai, Andrew M and Tong, Simon and Lepikhin, Dmitry and Xu, Yuanzhong and Krikun, Maxim and Zhou, Yanqi and Yu, Adams Wei and Firat, Orhan and others},
  booktitle={International Conference on Machine Learning},
  pages={5547--5569},
  year={2022},
  organization={PMLR}
}

@article{wenzek2019ccnet,
  title={CCNet: Extracting high quality monolingual datasets from web crawl data},
  author={Wenzek, Guillaume and Lachaux, Marie-Anne and Conneau, Alexis and Chaudhary, Vishrav and Guzm{\'a}n, Francisco and Joulin, Armand and Grave, Edouard},
  journal={arXiv preprint arXiv:1911.00359},
  year={2019}
}

@article{laurenccon2022bigscience,
  title={The BigScience ROOTS Corpus: A 1.6TB Composite Multilingual Dataset},
  author={Lauren{\c{c}}on, Hugo and Saulnier, Lucile and Wang, Thomas and Akiki, Christopher and Villanova del Moral, Albert and Le Scao, Teven and Von Werra, Leandro and Mou, Chenghao and Gonz{\'a}lez Ponferrada, Eduardo and Nguyen, Huu and Frohberg, J{\"o}rg and {\v{S}}a{\v{s}}ko, Mario and Lhoest, Quentin and McMillan-Major, Angelina and Dupont, Gerard and Biderman, Stella and Rogers, Anna and Ben Allal, Loubna and De Toni, Francesco and Pistilli, Giada and Nguyen, Olivier and Nikpoor, Somaieh and Masoud, Maraim and Colombo, Pierre and de la Rosa, Javier and Villegas, Paulo and Thrush, Tristan and Longpre, Shayne and Nagel, Sebastian and Weber, Leon and Mu{\~n}oz, Manuel and Zhu, Jian and Van Strien, Daniel and Alyafeai, Zaid and Almubarak, Khalid and Vu, Minh Chien and Gonzalez-Dios, Itziar and Soroa, Aitor and Lo, Kyle and Dey, Manan and Suarez, Pedro Ortiz and Gokaslan, Aaron and Bose, Shamik and Adelani, David and Phan, Long and Tran, Hieu and Yu, Ian and Pai, Suhas and Chim, Jenny and Lepercq, Violette and Ili{\'c}, Suzana and Mitchell, Margaret and Luccioni, Sasha Alexandra and Jernite, Yacine},
  journal={Advances in Neural Information Processing Systems},
  volume={35},
  pages={31809--31826},
  year={2022}
}

@inproceedings{yang2025novel,
  title={A novel diffusion model for pairwise geoscience data generation with unbalanced training dataset},
  author={Yang, Junhuan and Zhang, Yuzhou and Sheng, Yi and Lin, Youzuo and Yang, Lei},
  booktitle={Proceedings of the AAAI Conference on Artificial Intelligence},
  volume={39},
  number={20},
  pages={21965--21973},
  year={2025}
}

@article{yang2026digit,
  title={DiGiT: A Diffusion-based modular geophysical toolkit for on-device multi-modal data generation},
  author={Yang, Junhuan and Zhang, Yuzhou and Sheng, Yi and Lin, Youzuo and Jiang, Weiwen and Yang, Lei},
  journal={ACM Transactions on Embedded Computing Systems},
  volume={25},
  number={3},
  pages={1--26},
  year={2026},
  publisher={ACM New York, NY}
}

@article{brown2020language,
  title={Language models are few-shot learners},
  author={Brown, Tom and Mann, Benjamin and Ryder, Nick and Subbiah, Melanie and Kaplan, Jared D and Dhariwal, Prafulla and Neelakantan, Arvind and Shyam, Pranav and Sastry, Girish and Askell, Amanda and Sandhini Agarwal and  Ariel Herbert-Voss and Gretchen Krueger and  Tom Henighan and  Rewon Child and  Aditya Ramesh and  Daniel M. Ziegler and Jeffrey Wu and  Clemens Winter and  Christopher Hesse and  Mark Chen and  Eric Sigler and  Mateusz Litwin and  Scott Gray and Benjamin Chess and  Jack Clark and  Christopher Berner and  Sam McCandlish and Alec Radford and  Ilya Sutskever and  Dario Amodei},
  journal={Advances in neural information processing systems},
  volume={33},
  pages={1877--1901},
  year={2020}
}

@inproceedings{kingma2014adam,
  title={Adam: A method for stochastic optimization},
  author={Kingma, Diederik P and Ba, Jimmy},
  booktitle={International Conference on Learning Representations},
  year={2015}
}

@inproceedings{sagawa2019distributionally,
  title={Distributionally Robust Neural Networks},
  author={Sagawa, Shiori and Koh, Pang Wei and Hashimoto, Tatsunori B and Liang, Percy},
  booktitle={International Conference on Learning Representations (ICLR)},
  year={2019}
}

@article{grazzi2022bilevel,
  title={Bilevel Optimization with a Lower-level Contraction: Optimal Sample Complexity without Warm-Start},
  author={Grazzi, Riccardo and Pontil, Massimiliano and Salzo, Saverio},
  journal={arXiv preprint arXiv:2202.03397},
  year={2022}
}

@article{borsos2020coresets,
  title={Coresets via bilevel optimization for continual learning and streaming},
  author={Borsos, Zal{\'a}n and Mutny, Mojmir and Krause, Andreas},
  journal={Advances in Neural Information Processing Systems},
  volume={33},
  pages={14879--14890},
  year={2020}
}

@inproceedings{kwon2023fully,
  title={A fully first-order method for stochastic bilevel optimization},
  author={Kwon, Jeongyeol and Kwon, Dohyun and Wright, Stephen and Nowak, Robert D},
  booktitle={International Conference on Machine Learning},
  pages={18083--18113},
  year={2023},
  organization={PMLR}
}

@inproceedings{
gong2024a,
title={A Nearly Optimal Single Loop Algorithm for Stochastic Bilevel Optimization under Unbounded Smoothness},
author={Xiaochuan Gong and Jie Hao and Mingrui Liu},
booktitle={Forty-first International Conference on Machine Learning},
year={2024},
}

@article{xie2023doremi,
  title={DoReMi: Optimizing Data Mixtures Speeds Up Language Model Pretraining},
  author={Xie, Sang Michael and Pham, Hieu and Dong, Xuanyi and Du, Nan and Liu, Hanxiao and Lu, Yifeng and Liang, Percy and Le, Quoc V and Ma, Tengyu and Yu, Adams Wei},
  journal={Advances in Neural Information Processing Systems},
  year={2023}
}

@article{oren2019distributionally,
  title={Distributionally robust language modeling},
  author={Oren, Yonatan and Sagawa, Shiori and Hashimoto, Tatsunori B and Liang, Percy},
  journal={arXiv preprint arXiv:1909.02060},
  year={2019}
}

@inproceedings{
hao2024bilevel,
title={Bilevel Optimization under Unbounded Smoothness: A New Algorithm and Convergence Analysis},
author={Jie Hao and Xiaochuan Gong and Mingrui Liu},
booktitle={The Twelfth International Conference on Learning Representations},
year={2024},
}

@article{hao2023bilevel1,
  title={Bilevel Coreset Selection in Continual Learning: A New Formulation and Algorithm},
  author={Hao, Jie and Ji, Kaiyi and Liu, Mingrui},
  journal={Advances in Neural Information Processing Systems},
  volume={36},
  year={2023}
}

@article{hu2022lora,
  title={Lora: Low-rank adaptation of large language models.},
  author={Hu, Edward J and Shen, Yelong and Wallis, Phillip and Allen-Zhu, Zeyuan and Li, Yuanzhi and Wang, Shean and Wang, Lu and Chen, Weizhu and others},
  journal={ICLR},
  volume={1},
  number={2},
  pages={3},
  year={2022}
}

@article{hong2023two,
  title={A two-timescale stochastic algorithm framework for bilevel optimization: Complexity analysis and application to actor-critic},
  author={Hong, Mingyi and Wai, Hoi-To and Wang, Zhaoran and Yang, Zhuoran},
  journal={SIAM Journal on Optimization},
  volume={33},
  number={1},
  pages={147--180},
  year={2023},
  publisher={SIAM}
}

@book{dempe2002foundations,
  title={Foundations of bilevel programming},
  author={Dempe, Stephan},
  year={2002},
  publisher={Springer Science \& Business Media}
}

@inproceedings{chen2023bilevel1,
  title={On Finding Small Hyper-Gradients in Bilevel Optimization: Hardness Results and Improved Analysis},
  author={Chen, Lesi and Xu, Jing and Zhang, Jingzhao},
  booktitle={Proceedings of Thirty Seventh Conference on Learning Theory},
  pages={947--980},
  year={2024},
  volume={247},
  series={Proceedings of Machine Learning Research},
  publisher={PMLR},
  url={https://proceedings.mlr.press/v247/chen24a.html}
}

@article{bracken1973mathematical,
  title={Mathematical programs with optimization problems in the constraints},
  author={Bracken, Jerome and McGill, James T},
  journal={Operations Research},
  volume={21},
  number={1},
  pages={37--44},
  year={1973},
  publisher={INFORMS}
}

@article{dagreou2022framework,
  title={A framework for bilevel optimization that enables stochastic and global variance reduction algorithms},
  author={Dagr{\'e}ou, Mathieu and Ablin, Pierre and Vaiter, Samuel and Moreau, Thomas},
  journal={arXiv preprint arXiv:2201.13409},
  year={2022}
}

@inproceedings{franceschi2018bilevel,
  title={Bilevel programming for hyperparameter optimization and meta-learning},
  author={Franceschi, Luca and Frasconi, Paolo and Salzo, Saverio and Grazzi, Riccardo and Pontil, Massimiliano},
  booktitle={International Conference on Machine Learning},
  pages={1568--1577},
  year={2018},
  organization={PMLR}
}

@inproceedings{zhou2022probabilistic,
  title={Probabilistic Bilevel Coreset Selection},
  author={Zhou, Xiao and Pi, Renjie and Zhang, Weizhong and Lin, Yong and Chen, Zonghao and Zhang, Tong},
  booktitle={International Conference on Machine Learning},
  pages={27287--27302},
  year={2022},
  organization={PMLR}
}

@inproceedings{finn2017model,
  title={Model-agnostic meta-learning for fast adaptation of deep networks},
  author={Finn, Chelsea and Abbeel, Pieter and Levine, Sergey},
  booktitle={International conference on machine learning},
  pages={1126--1135},
  year={2017},
  organization={PMLR}
}

@inproceedings{xia2024less,
  title={LESS: selecting influential data for targeted instruction tuning},
  author={Xia, Mengzhou and Malladi, Sadhika and Gururangan, Suchin and Arora, Sanjeev and Chen, Danqi},
  booktitle={Proceedings of the 41st International Conference on Machine Learning},
  pages={54104--54132},
  year={2024}
}

@article{xie2023data,
  title={Data selection for language models via importance resampling},
  author={Xie, Sang Michael and Santurkar, Shibani and Ma, Tengyu and Liang, Percy S},
  journal={Advances in Neural Information Processing Systems},
  volume={36},
  pages={34201--34227},
  year={2023}
}

@inproceedings{ji2021bilevel,
  title={Bilevel optimization: Convergence analysis and enhanced design},
  author={Ji, Kaiyi and Yang, Junjie and Liang, Yingbin},
  booktitle={International conference on machine learning},
  pages={4882--4892},
  year={2021},
  organization={PMLR}
}

@article{ghadimi2018approximation,
  title={Approximation methods for bilevel programming},
  author={Ghadimi, Saeed and Wang, Mengdi},
  journal={arXiv preprint arXiv:1802.02246},
  year={2018}
}

@article{gao2020pile,
  title={The pile: An 800gb dataset of diverse text for language modeling},
  author={Gao, Leo and Biderman, Stella and Black, Sid and Golding, Laurence and Hoppe, Travis and Foster, Charles and Phang, Jason and He, Horace and Thite, Anish and Nabeshima, Noa and  Shawn Presser and Connor Leahy},
  journal={arXiv preprint arXiv:2101.00027},
  year={2020}
}

@article{conneau2019cross,
  title={Cross-lingual language model pretraining},
  author={Conneau, Alexis and Lample, Guillaume},
  journal={Advances in neural information processing systems},
  volume={32},
  year={2019}
}

@inproceedings{koh2017understanding,
  title={Understanding black-box predictions via influence functions},
  author={Koh, Pang Wei and Liang, Percy},
  booktitle={International conference on machine learning},
  pages={1885--1894},
  year={2017},
  organization={PMLR}
}

@article{hampel1974influence,
  title={The influence curve and its role in robust estimation},
  author={Hampel, Frank R},
  journal={Journal of the american statistical association},
  volume={69},
  number={346},
  pages={383--393},
  year={1974},
  publisher={Taylor \& Francis}
}

@article{fan2023doge,
  title={Doge: Domain reweighting with generalization estimation},
  author={Fan, Simin and Pagliardini, Matteo and Jaggi, Martin},
  journal={arXiv preprint arXiv:2310.15393},
  year={2023}
}

@inproceedings{somayajula2023bi,
  title={Bi-level Finetuning with Task-dependent Similarity Structure for Low-resource Training},
  author={Somayajula, Sai Ashish and Jin, Lifeng and Song, Linfeng and Mi, Haitao and Yu, Dong},
  booktitle={Findings of the Association for Computational Linguistics: ACL 2023},
  pages={8569--8588},
  year={2023}
}

@article{pan2024scalebio,
  title={ScaleBiO: Scalable Bilevel Optimization for LLM Data Reweighting},
  author={Pan, Rui and Zhang, Jipeng and Pan, Xingyuan and Pi, Renjie and Wang, Xiaoyu and Zhang, Tong},
  journal={arXiv preprint arXiv:2406.19976},
  year={2024}
}

@inproceedings{grangier2023bilevel,
  title={Bilevel Optimization to Learn Training Distributions for Language Modeling under Domain Shift},
  author={Grangier, David and Ablin, Pierre and Hannun, Awni},
  booktitle={NeurIPS 2023 Workshop on Distribution Shifts: New Frontiers with Foundation Models},
  year={2023},
  url={https://openreview.net/forum?id=D67r01BYYP},
  note={Extended version available as arXiv:2311.11973}
}

@article{chen2023skill,
  title={Skill-it! a data-driven skills framework for understanding and training language models},
  author={Chen, Mayee and Roberts, Nicholas and Bhatia, Kush and Wang, Jue and Zhang, Ce and Sala, Frederic and R{\'e}, Christopher},
  journal={Advances in Neural Information Processing Systems},
  volume={36},
  pages={36000--36040},
  year={2023}
}

@article{cook1977detection,
  title={Detection of influential observation in linear regression},
  author={Cook, R Dennis},
  journal={Technometrics},
  volume={19},
  number={1},
  pages={15--18},
  year={1977},
  publisher={Taylor \& Francis}
}

@book{ling1984residuals,
  title={Residuals and Influence in Regression},
  author={Cook, R Dennis and Weisberg, Sanford},
  year={1982},
  publisher={Chapman and Hall}
}

@article{maini2024rephrasing,
  title={Rephrasing the web: A recipe for compute and data-efficient language modeling},
  author={Maini, Pratyush and Seto, Skyler and Bai, He and Grangier, David and Zhang, Yizhe and Jaitly, Navdeep},
  journal={arXiv preprint arXiv:2401.16380},
  year={2024}
}

@article{sorscher2022beyond,
  title={Beyond neural scaling laws: beating power law scaling via data pruning},
  author={Sorscher, Ben and Geirhos, Robert and Shekhar, Shashank and Ganguli, Surya and Morcos, Ari},
  journal={Advances in Neural Information Processing Systems},
  volume={35},
  pages={19523--19536},
  year={2022}
}

@article{tirumala2023d4,
  title={D4: Improving llm pretraining via document de-duplication and diversification},
  author={Tirumala, Kushal and Simig, Daniel and Aghajanyan, Armen and Morcos, Ari},
  journal={Advances in Neural Information Processing Systems},
  volume={36},
  pages={53983--53995},
  year={2023}
}

@article{penedo2023refinedweb,
  title={The RefinedWeb dataset for Falcon LLM: outperforming curated corpora with web data, and web data only},
  author={Penedo, Guilherme and Malartic, Quentin and Hesslow, Daniel and Cojocaru, Ruxandra and Cappelli, Alessandro and Alobeidli, Hamza and Pannier, Baptiste and Almazrouei, Ebtesam and Launay, Julien},
  journal={arXiv preprint arXiv:2306.01116},
  year={2023}
}

@article{abbas2023semdedup,
  title={Semdedup: Data-efficient learning at web-scale through semantic deduplication},
  author={Abbas, Amro and Tirumala, Kushal and Simig, D{\'a}niel and Ganguli, Surya and Morcos, Ari S},
  journal={arXiv preprint arXiv:2303.09540},
  year={2023}
}

@article{lee2021deduplicating,
  title={Deduplicating training data makes language models better},
  author={Lee, Katherine and Ippolito, Daphne and Nystrom, Andrew and Zhang, Chiyuan and Eck, Douglas and Callison-Burch, Chris and Carlini, Nicholas},
  journal={arXiv preprint arXiv:2107.06499},
  year={2021}
}

@inproceedings{wettig2024qurating,
  title={Qurating: Selecting high-quality data for training language models},
  author={Wettig, Alexander and Gupta, Aatmik and Malik, Saumya and Chen, Danqi},
  booktitle={Forty-first International Conference on Machine Learning},
  year={2024}
}

@inproceedings{albalak2023efficient,
  title={Efficient online data mixing for language model pre-training},
  author={Albalak, Alon and Pan, Liangming and Raffel, Colin and Wang, William Yang},
  booktitle={R0-FoMo: Robustness of Few-shot and Zero-shot Learning in Large Foundation Models},
  year={2023}
}

@article{xia2023sheared,
  title={Sheared LLaMA: Accelerating Language Model Pre-training via Structured Pruning},
  author={Xia, Mengzhou and Gao, Tianyu and Zeng, Zhiyuan and Chen, Danqi},
  journal={arXiv preprint arXiv:2310.06694},
  year={2023}
}

@article{yu2024mates,
  title={MATES: Model-Aware Data Selection for Efficient Pretraining with Data Influence Models},
  author={Yu, Zichun and Das, Spandan and Xiong, Chenyan},
  journal={arXiv preprint arXiv:2406.06046},
  year={2024}
}

@article{engstrom2024dsdm,
  title={Dsdm: Model-aware dataset selection with datamodels},
  author={Engstrom, Logan and Feldmann, Axel and Madry, Aleksander},
  journal={arXiv preprint arXiv:2401.12926},
  year={2024}
}

@article{albalak2024survey,
  title={A survey on data selection for language models},
  author={Albalak, Alon and Elazar, Yanai and Xie, Sang Michael and Longpre, Shayne and Lambert, Nathan and Wang, Xinyi and Muennighoff, Niklas and Hou, Bairu and Pan, Liangming and Jeong, Haewon and others},
  journal={arXiv preprint arXiv:2402.16827},
  year={2024}
}

@article{li2024datacomp,
  title={Datacomp-lm: In search of the next generation of training sets for language models},
  author={Li, Jeffrey and Fang, Alex and Smyrnis, Georgios and Ivgi, Maor and Jordan, Matt and Gadre, Samir and Bansal, Hritik and Guha, Etash and Keh, Sedrick and Arora, Kushal and others},
  journal={arXiv preprint arXiv:2406.11794},
  year={2024}
}

@inproceedings{park2023trak,
  title={TRAK: attributing model behavior at scale},
  author={Park, Sung Min and Georgiev, Kristian and Ilyas, Andrew and Leclerc, Guillaume and M{\k{a}}dry, Aleksander},
  booktitle={Proceedings of the 40th International Conference on Machine Learning},
  pages={27074--27113},
  year={2023}
}

@article{rae2021scaling,
  title={Scaling language models: Methods, analysis \& insights from training gopher},
  author={Rae, Jack W and Borgeaud, Sebastian and Cai, Trevor and Millican, Katie and Hoffmann, Jordan and Song, Francis and Aslanides, John and Henderson, Sarah and Ring, Roman and Young, Susannah and others},
  journal={arXiv preprint arXiv:2112.11446},
  year={2021}
}

@article{welbl2017crowdsourcing,
  title={Crowdsourcing multiple choice science questions},
  author={Welbl, Johannes and Liu, Nelson F and Gardner, Matt},
  journal={arXiv preprint arXiv:1707.06209},
  year={2017}
}

@article{clark2018think,
  title={Think you have solved question answering? try arc, the ai2 reasoning challenge},
  author={Clark, Peter and Cowhey, Isaac and Etzioni, Oren and Khot, Tushar and Sabharwal, Ashish and Schoenick, Carissa and Tafjord, Oyvind},
  journal={arXiv preprint arXiv:1803.05457},
  year={2018}
}

@article{liu2020logiqa,
  title={Logiqa: A challenge dataset for machine reading comprehension with logical reasoning},
  author={Liu, Jian and Cui, Leyang and Liu, Hanmeng and Huang, Dandan and Wang, Yile and Zhang, Yue},
  journal={arXiv preprint arXiv:2007.08124},
  year={2020}
}

@article{mihaylov2018can,
  title={Can a suit of armor conduct electricity? a new dataset for open book question answering},
  author={Mihaylov, Todor and Clark, Peter and Khot, Tushar and Sabharwal, Ashish},
  journal={arXiv preprint arXiv:1809.02789},
  year={2018}
}

@article{clark2019boolq,
  title={BoolQ: Exploring the surprising difficulty of natural yes/no questions},
  author={Clark, Christopher and Lee, Kenton and Chang, Ming-Wei and Kwiatkowski, Tom and Collins, Michael and Toutanova, Kristina},
  journal={arXiv preprint arXiv:1905.10044},
  year={2019}
}

@article{zellers2019hellaswag,
  title={Hellaswag: Can a machine really finish your sentence?},
  author={Zellers, Rowan and Holtzman, Ari and Bisk, Yonatan and Farhadi, Ali and Choi, Yejin},
  journal={arXiv preprint arXiv:1905.07830},
  year={2019}
}

@inproceedings{bisk2020piqa,
  title={Piqa: Reasoning about physical commonsense in natural language},
  author={Bisk, Yonatan and Zellers, Rowan and Gao, Jianfeng and Choi, Yejin and others},
  booktitle={Proceedings of the AAAI conference on artificial intelligence},
  volume={34},
  pages={7432--7439},
  year={2020}
}

@article{sakaguchi2021winogrande,
  title={Winogrande: An adversarial winograd schema challenge at scale},
  author={Sakaguchi, Keisuke and Bras, Ronan Le and Bhagavatula, Chandra and Choi, Yejin},
  journal={Communications of the ACM},
  volume={64},
  number={9},
  pages={99--106},
  year={2021},
  publisher={ACM New York, NY, USA}
}

@article{rajpurkar2016squad,
  title={SQuAD: 100,000+ Questions for Machine Comprehension of Text},
  author={Rajpurkar, Pranav and Zhang, Jian and Lopyrev, Konstantin and Liang, Percy},
  journal={arXiv preprint arXiv:1606.05250},
  year={2016}
}

@misc{gao2021framework,
  title={A framework for few-shot language model evaluation},
  author={Gao, Leo and Tow, Jonathan and Biderman, Stella and Black, Sid and DiPofi, Anthony and Foster, Charles and Golding, Laurence and Hsu, Jeffrey and McDonell, Kyle and Muennighoff, Niklas and Phang, Jason and Reynolds, Laria and Tang, Eric and Thite, Anish and Wang, Ben and Wang, Kevin and Zou, Andy},
  year={2021},
  publisher={Zenodo},
  version={v0.0.1},
  doi={10.5281/zenodo.5371629},
  url={https://doi.org/10.5281/zenodo.5371629}
}

@misc{openai2024terms,
  author    = "{OpenAI}",
  title     = "{OpenAI Terms of Service}",
  year      = {2024},
  url       = {https://openai.com/terms},
  note      = "Accessed: Jan 30, 2025"
}

@misc{google2024geminiterms,
  author    = "{Google}",
  title     = "{Gemini API Additional Terms of Service}",
  year      = {2024},
  url       = {https://ai.google.dev/gemini-api/terms},
  note      = "Accessed: January 30, 2025"
}

@article{pan2025alinfik,
  title={ALinFiK: Learning to Approximate Linearized Future Influence Kernel for Scalable Third-Party LLM Data Valuation},
  author={Pan, Yanzhou and Lin, Huawei and Ran, Yide and Chen, Jiamin and Yu, Xiaodong and Zhao, Weijie and Zhang, Denghui and Xu, Zhaozhuo},
  journal={arXiv preprint arXiv:2503.01052},
  year={2025}
}

@article{lin2024token,
  title={Token-wise influential training data retrieval for large language models},
  author={Lin, Huawei and Long, Jikai and Xu, Zhaozhuo and Zhao, Weijie},
  journal={arXiv preprint arXiv:2405.11724},
  year={2024}
}

@inproceedings{zhuang2025meta,
  title={Meta-rater: A multi-dimensional data selection method for pre-training language models},
  author={Zhuang, Xinlin and Peng, Jiahui and Ma, Ren and Wang, Yinfan and Bai, Tianyi and Wei, Xingjian and Jiantao, Qiu and Zhang, Chi and Qian, Ying and He, Conghui},
  booktitle={Proceedings of the 63rd Annual Meeting of the Association for Computational Linguistics (Volume 1: Long Papers)},
  pages={10856--10896},
  year={2025}
}

@article{zhang2409harnessing,
  title={Harnessing Diversity for Important Data Selection in Pretraining Large Language Models},
  author={Zhang, C and Zhong, H and Zhang, K and Chai, C and Wang, R and Zhuang, X and Bai, T and Qiu, J and Cao, L and Fan, J and others},
  journal={arXiv preprint arXiv:2409.16986},
year = {2024}
}

@misc{slimpajama6b,
  title        = {SlimPajama-6B},
  author       = {DKYoon},
  year         = {2023},
  howpublished = {\texttt{HuggingFace Hub}},
  note         = {\url{https://huggingface.co/datasets/DKYoon/SlimPajama-6B}}
}

@inproceedings{kwon2024datainf,
  title={DataInf: Efficiently Estimating Data Influence in LoRA-tuned LLMs and Diffusion Models},
  author={Kwon, Yongchan and Wu, Eric and Wu, Kevin and Zou, James Y},
  booktitle={International Conference on Learning Representations},
  year={2024}
}

@article{shin2024dash,
  title={Dash: Warm-starting neural network training in stationary settings without loss of plasticity},
  author={Shin, Baekrok and Oh, Junsoo and Cho, Hanseul and Yun, Chulhee},
  journal={Advances in Neural Information Processing Systems},
  volume={37},
  pages={43300--43340},
  year={2024}
}

@article{salehi2025bilevel,
  title={Bilevel Learning via Inexact Stochastic Gradient Descent},
  author={Salehi, Mohammad Sadegh and Mukherjee, Subhadip and Roberts, Lindon and Ehrhardt, Matthias J},
  journal={arXiv preprint arXiv:2511.06774},
  year={2025}
}

@inproceedings{gong2026adaptive,
  title={Adaptive Algorithms with Sharp Convergence Rates for Stochastic Hierarchical Optimization},
  author={Gong, Xiaochuan and Hao, Jie and Liu, Mingrui},
  booktitle={Advances in Neural Information Processing Systems},
  volume={38},
  year={2025}
}

@article{wu2026bilevel,
  title={Bilevel Optimization with Lower-Level Uniform Convexity: Theory and Algorithm},
  author={Wu, Yuman and Gong, Xiaochuan and Hao, Jie and Liu, Mingrui},
  journal={International Conference on Learning Representations},
  year={2026}
}

@article{gong2024accelerated,
  title={An accelerated algorithm for stochastic bilevel optimization under unbounded smoothness},
  author={Gong, Xiaochuan and Hao, Jie and Liu, Mingrui},
  journal={Advances in Neural Information Processing Systems},
  volume={37},
  pages={78201--78243},
  year={2024}
}
\bibliographystyle{arxiv}

%%%%%%%%%%%%%%%%%%%%%%%%%%%%%%%%%%%%%%%%%%%%%%%%%%%%%%%%%%%%%%%%%%%%%%%%%%%%%%%
%%%%%%%%%%%%%%%%%%%%%%%%%%%%%%%%%%%%%%%%%%%%%%%%%%%%%%%%%%%%%%%%%%%%%%%%%%%%%%%
% APPENDIX
%%%%%%%%%%%%%%%%%%%%%%%%%%%%%%%%%%%%%%%%%%%%%%%%%%%%%%%%%%%%%%%%%%%%%%%%%%%%%%%
%%%%%%%%%%%%%%%%%%%%%%%%%%%%%%%%%%%%%%%%%%%%%%%%%%%%%%%%%%%%%%%%%%%%%%%%%%%%%%%
\newpage
\appendix
\onecolumn
\section{Convergence Analysis and Practical Design Choices}
\label{app:analysis}

Classical bilevel convergence analyses typically rely on a strongly-convex lower-level problem, so that the lower-level solution mapping is well-defined and smooth. This assumption also underlies many recent single-loop bilevel methods that update the upper-level and lower-level variables with the same frequency~\cite{dagreou2022framework}. However, this formulation is not necessarily suitable for BLISS. In our setting, the lower-level corresponds to training a proxy language model under score-induced sample weights, which may not be strongly convex.

%is nonconvex rather than strongly convex. Therefore, the exact lower-level minimizer is neither easy to characterize nor necessary for describing the actual BLISS update.

To characterize the update rule of BLISS, we adopt a finite-horizon view and define the lower level through \(N\) steps of stochastic gradient descent (SGD) on the proxy model:
\begin{equation}\label{eq:fh_bliss}
\begin{aligned}
\min_{\theta_s}\quad \Phi_{N}(\theta_s)
&:= 
\mathbb{E}_{\zeta\sim\mathcal D_{\mathrm{val}}}
\big[F(\theta_p^{N}(\theta_s);\zeta)\big],
\qquad \text{(UL)}\\
\text{s.t.}\quad
% \theta_p^{N}(\theta_s)
% &= \mathrm{SGD}\big(G(\theta_p, \theta_s; \xi)\big),
\theta_p^{N}(\theta_s) &= \texttt{SGD}(\theta_p, \nabla_{ \theta_p} G(\theta_p, \theta_s; \xi), \eta_1, N)
\qquad \xi\sim\mathcal D_{\mathrm{tr}} .
\qquad \text{(LL)}
\end{aligned}
\end{equation}
Here, given \(\theta_s\), the proxy model is trained for \(N\) SGD steps with step size $\eta_1$, and the upper-level objective is evaluated on the resulting finite-horizon proxy model. This surrogate objective \(\Phi_N\) is faithful to BLISS, since each upper-level update interacts with a proxy model obtained under a fixed computational budget, rather than with an idealized exact lower-level optimum.

We next introduce a finite-horizon approximation assumption, which plays the same role as the lower-level accuracy condition in standard inexact bilevel analysis. Let \(\theta_p^*(\theta_s)\) denote the unique optimal proxy-training target induced by the score model, and define the \(N\)-step lower-level approximation error at outer iteration $t$ as
\[
\delta_t^2
:=
\sup_{\theta_s^t}
\mathbb{E}
\left[
\left\|
\theta_p^N(\theta_s^t)
-
\theta_p^*(\theta_s^t)
\right\|^2
\right].
\]
We follow the assumptions in~\citep{salehi2025bilevel} to derive the convergence rate, including function smoothness and differentiable Assumption 3.1, and step size Assumption 3.3 in~\citep{salehi2025bilevel}.
Let \(\nabla\Phi_N(\theta_s)\) be the upper-level stochastic gradient used by BLISS and let \(\Phi(\theta_s)\) denote the ideal target objective. As in~\cite{salehi2025bilevel}, we assume the upper-level stochastic gradient
\[
\nabla\Phi_N(\theta_s^t)
=
\nabla \hat \Phi(\theta_s^t)
+
e_t(\theta_s^t),
\]
where $\mathbb{E}[\nabla \hat \Phi(\theta_s^t)]=\nabla \Phi(\theta_s^t)$, and \(e_t(\theta_s^t)\) is the finite-horizon gradient error satisfying
\[
\mathbb{E}\|e_t(\theta_s^t)\|^2
\le
\delta_t^2 .
\]
Thus, \(\delta_t\) measures the maximal finite-horizon lower-level error, while \(e_t(\theta_s^t)\) denotes the corresponding inexact hypergradient error. In each iteration $t$, the upper level variable is updated by one-step inexact stochastic gradient descent over $\theta_s$ (i.e., $\theta_s^{t+1} = \theta_s^{t} - \alpha_t \nabla\Phi_N(\theta_s^t)$) while the lower-level is updated by $N$ steps of stochastic gradient descent over $\theta_p$.

Therefore, we can get the convergence theorem as in Theorem~3.10 of \citep{salehi2025bilevel},
the BLISS iterates satisfy
\[
\min_{0\le t\le T-1}
\mathbb{E}
\left[
\|\nabla \Phi(\theta_s^t)\|^2
\right]
\le
\frac{C_1}{T\alpha_T}
+
\frac{C_2}{T\alpha_T}
\sum_{t=0}^{T-1}
\alpha_t \delta_{t}^2,
\]
where $C_1,C_2>0$ are constants independent of $T$. Following the setting in Corollary 3.13 of \citet{salehi2025bilevel}, assuming the error $\delta_t=O(t^{-p})$ with $p\ge 1/4$, we can choose the step size $\alpha_t= O(t^{-q})$ with $q \downarrow 1/2$ (choose $q$ arbitrarily close $1/2$ from above), and then we can achieve the convergence rate of $O(T^{-1/4})$ of the gradient norm.

% Under standard smoothness and lower-boundedness assumptions on \(\Phi\), the standard inexact stochastic-gradient analysis implies that, with the upper-level learning-rate choice \(\eta_t=O(t^{-q})\) with  $q \uparrow 1/2$ (chose $q$ arbitrarily close $1/2$ from above),
% \[
% \frac{1}{T}\sum_{t=0}^{T-1}
% \mathbb{E}
% \|\nabla \Phi(\theta_s^t)\|
% \le
% O\left(
% T^{-1/4}
% +
% \delta_N
% \right)
% \]
% Therefore, BLISS achieves the standard stochastic nonconvex rate \(O(T^{-1/4})\) in expected gradient norm, up to the finite-horizon lower-level approximation error. If \(\delta_N=O(N^{-r})\) for some \(r>0\), the bound becomes
% \[
% O\left(T^{-1/4}+N^{-r}\right).
% \]
% Let $r \ge 1/4$ , the convergence rate achieves $O(T^{-1/4})$.

In practice, the lower-level step budget $N$ is chosen empirically to ensure that the surrogate lower-level solution is sufficiently close to the target solution while keeping the computation affordable. For example, in our 31M proxy setting, we find that 1 lower-level step already gives a reasonably good approximation, whereas in the 160M proxy setting, we use 5 lower-level steps. This suggests that the choice of $N$ depends on the proxy size and is tuned to keep the finite-horizon approximation error sufficiently small in practice.
For schedule of learning rate, we apply cosine learning decay to the upper level learning rate $\alpha$.

For choice of the selection ratio, this hyperparameter is decided by the computational budget, since the selected data is used to pretrain the target LLM, which takes over most of the computational resources. This setting also follows form the prior work MATE \citep{yu2024mates}  and DsDm \citep{engstrom2024dsdm}, the computational budget limits larger choices of this hyperparameter. 

\section{Details of Model Settings} \label{sec:model_setting}

The proxy model and score model serve different purposes: the proxy model acts as a surrogate for the LLM and is trained for next-token prediction, while the score model functions as a regression model that maps individual samples to their corresponding influence scores. To transform the proxy model into the score model, we modify its architecture by replacing the final \texttt{Linear} layer with an \texttt{AdaptiveAvgPool} layer, followed by a \texttt{Linear} layer and a \texttt{Sigmoid} activation. Specifically, given the output from the preceding transformer blocks with dimension $[\texttt{Batch}, \texttt{token\_size}, \texttt{Emb\_size}]$, the \texttt{AdaptiveAvgPool} layer computes the average embedding feature across tokens. The \texttt{Linear} layer then maps the pooled token representations to a single-dimensional output, which is subsequently passed through a \texttt{Sigmoid} activation to produce an influence score within the range $(0,1)$. In contrast, the proxy model’s final \texttt{Linear} layer maps features from previous layers to the vocabulary dimension for token prediction. 

\section{Implementation Details in LLaMA Experiment}\label{sec:llama}

\paragraph{Model Setup}
In LLaMA setting, the target model is LLaMA-0.5B, and the proxy/score model is LLaMA-134M. They are warmed up under the same process as Pythia setting.

\paragraph{Training Details of Proxy/Score Model}
There is a little difference in how we deal with the proxy model in LLaMA setting compared to Pythia setting. In addition to resetting the proxy model (LLaMA-134M) at the beginning of each round, we reset it to the initial state every 50 steps of the update of the score model. We distill the target model into the proxy model by minimizing the KL divergence for 240 steps. Then the checkpoint of the proxy model is saved as ``initial" state. Since periodically resetting the proxy model ensures a close alignment between two models, we remove the KL divergence regularization term in the lower level loss function. To achieve a better lower-level solution, the proxy model executes 4 lower-level updates, each computed on a batch of 64 samples. After the score model is trained for 50 optimization steps, we reset the proxy model to the initial state.

\section{More Ablation Studies}\label{sec:mode_ablation}

In this section, we provide more ablation studies to verify the effectiveness of each component in our algorithm design. 

\begin{figure}[!ht]
    \centering
    \includegraphics[width=0.8\linewidth]{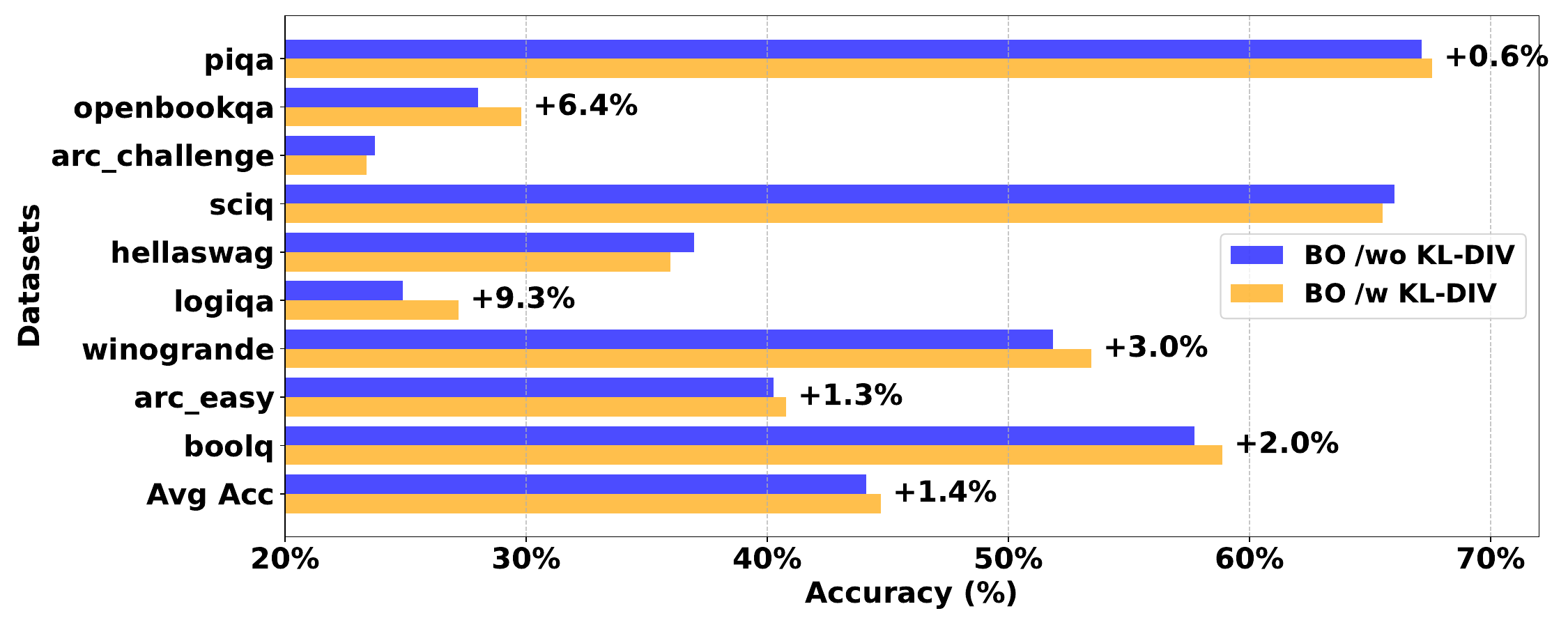}
    \caption{The performance comparison of bilevel optimization with/without KL divergence. The number on the bar indicate the accuracy improvement compared to the method without KL divergence.}
    \label{fig:training_without_kl2}
\end{figure}

% \vspace*{-0.05in}
\begin{table*}[!h]
    \centering
    \caption{Comparison of BLISS with different settings(without softmax and single level update) over multiple downstream datasets (410M model, 10B tokens) with 20k-step training. }
    % \vspace*{-0.09in}
    %The accuracy with standard error is reported based on the lm-evaluation-harness \cite{gao2021framework} implementation.}
    \resizebox{1\textwidth}{!}{
    \begin{tabular}{lcccccccccc}
    \toprule
\textbf{Methods} & \textbf{SciQ} & \textbf{ARC-E} & \textbf{ARC-C} & \textbf{LogiQA} & \textbf{OBQA} & \textbf{BoolQ} & \textbf{HellaSwag} & \textbf{PIQA} & \textbf{WinoGrande} & \textbf{Average} \\
  \midrule
Without softmax &63.5\scriptsize{(1.5)}	&41.0\scriptsize{(1.0)}	&22.4\scriptsize{(1.2)}	&25.7\scriptsize{(1.7)}	&30.0\scriptsize{(2.1)}	&52.8\scriptsize{(0.9)}	&38.8\scriptsize{(0.5)}	&67.4\scriptsize{(1.1)}	&51.0\scriptsize{(1.4)}	&43.6\scriptsize{(1.3)}
 \\
Single Level &64.4\scriptsize{(1.5)}	&42.3\scriptsize{(1.0)}	&22.2\scriptsize{(1.2)}	&24.1\scriptsize{(1.7)}	&30.6\scriptsize{(2.1)}	&55.0\scriptsize{(0.9)}	&39.7\scriptsize{(0.5)}	&67.1\scriptsize{(1.1)}	&52.1\scriptsize{(1.4)}	&44.2\scriptsize{(1.3)}\\

\multicolumn{1}{l}{BLISS} &65.5\scriptsize{(1.5)}	&40.8\scriptsize{(1.0)}	&23.4\scriptsize{(1.2)}	&27.2\scriptsize{(1.7)}	&29.8\scriptsize{(2.0)}	&58.9\scriptsize{(0.9)}	&36.0\scriptsize{(0.5)}	&67.6\scriptsize{(1.1)}	&53.4\scriptsize{(1.4)}	&44.7\scriptsize{(1.3)}

\\
  \bottomrule
\end{tabular}\label{tbl:single_level}}
% \vspace{-0.05in}
\end{table*}

\subsection{Softmax Reparametrization for Score Model's Output} \label{sec:softmax_score}

In our experiment, we apply a softmax function on all batch samples' score across GPUs to obtain the importance weights $P_i$. Note that the raw output of the score model is already within the range $(0,1)$, but we add another softmax function on top of it. We want to demonstrate the effectiveness of this softmax reparameterization. Intuitively, the main benefit is that it naturally amplifies important samples while downweighting less useful ones, improving the overall data selection process.

To assess the impact of the softmax reparameterization, we conduct an ablation experiment comparing two approaches: (i) naive weighting,  where the raw outputs of the score model are used directly as sample weights; (ii) softmax weighting, where  the softmax-transformed outputs of the score model determine the sample weights. The results, shown in Table \ref{tbl:single_level}, indicate that softmax weighting consistently outperforms naive weighting, leading to a 1.1\% improvement in average downstream accuracy. This demonstrates that softmax effectively enhances data selection by better distinguishing important samples.

% \subsection{Single-level vs. Bilevel Optimization} \label{sec:bilevel_opt}

% In bilevel algorithm, the hyper-gradient is essential for the update of upper level parameters. To verify the effectiveness of bilevel update of upper parameters, we compare bilevel update with simultaneous update, which update $\theta_s$ and $\theta_p$ together using both training and validation data for the lower-level objective. Specifically, the upper and lower levels are reduced to a single level problem: the upper-level and lower-level parameters are updated simultaneously on validation dataset and training dataset respectively. With the same number of training steps as bilevel training, the average accuracy of single level update degrades $0.5\%$ as shown in Table \ref{tbl:nosoftmax}.

\subsection{The Size of Proxy Model} \label{sec:model_size}

\begin{figure*}[t]
    \centering

    \begin{subfigure}{0.48\linewidth}
        \centering
        \includegraphics[width=0.48\linewidth]{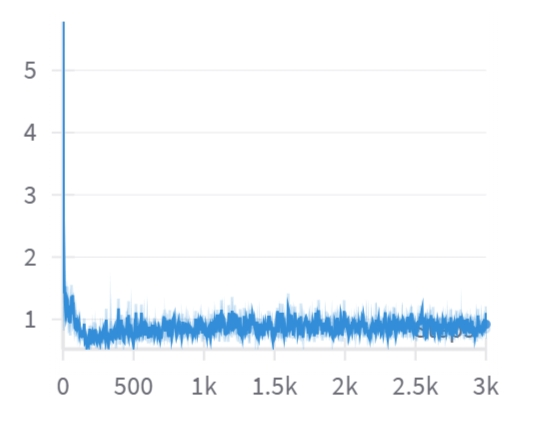}
        \includegraphics[width=0.48\linewidth]{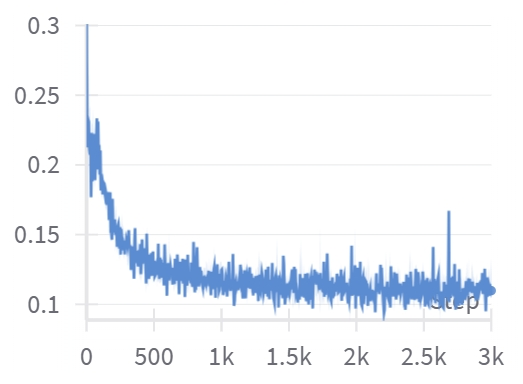}
        \caption{Proxy size 31M, target LLM 410M.}
        \label{fig:proxy_31m}
    \end{subfigure}
    \hfill
    \begin{subfigure}{0.48\linewidth}
        \centering
        \includegraphics[width=0.48\linewidth]{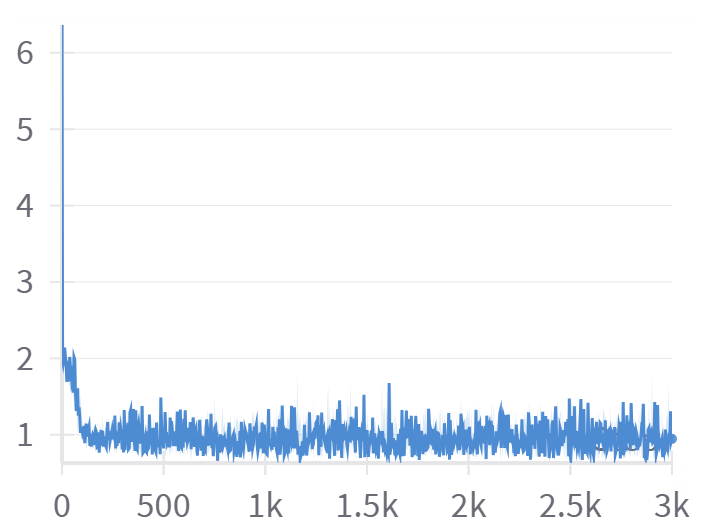}
        \includegraphics[width=0.48\linewidth]{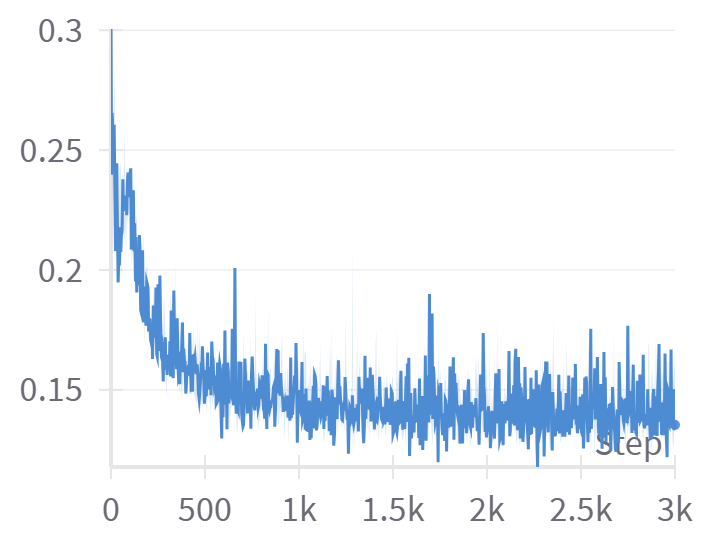}
        \caption{Proxy size 160M, target LLM 410M.}
        \label{fig:proxy_160m}
    \end{subfigure}

    \caption{Evolution of lower-level training loss and KL divergence for different proxy model sizes.}
    \label{fig:proxy_model_size}
\end{figure*}

We conduct experiments using two different sizes of proxy/score models (31M and 160M) for a 410M LLM. We observe that the KL divergence between the proxy and the LLM remains low for both sizes-0.15 for the 160M model and 0.10 for the 31M model. The corresponding learning curves are shown in Figure~\ref{fig:proxy_model_size}, which presents the results from round 2. The performance comparison of two sizes of proxy model is summarized in Table \ref{tab:proxy_model_size}. These findings suggest that even a small proxy model (31M) is sufficient to serve as an effective surrogate for the 410M LLM.

% We did experiments with two different sizes of proxy/score model (31M and 160M) for 410M LLM, and we found that the KL divergence for both sizes of the model are both small (0.15 for 160M model and 0.1 for 31M model). The learning curves are presented in Figure \ref{fig:proxy_model_size}, which is the result of round 2. Therefore, a small proxy model is sufficient to be the surrogate of the 410M LLM.
% which is similar to other proxy model-based data selection methods such as DsDm~\cite{engstrom2024dsdm}, DeReMi~\cite{xie2023doremi}, and MATES~\cite{yu2024mates}. For instance, DoReMi uses a 280M proxy model for a target LLM with 8B parameters, and MATES employs a 160M proxy model for a target LLM with 410M parameters. 

% \begin{figure*}[!h]
%     \centering
%         \subfigure[Training loss vs. steps]{\includegraphics[width=0.23\linewidth]{figures/train_loss_r1.png}}
%         \subfigure[KL divergence]{\includegraphics[width=0.23\linewidth]{figures/kl_div.png}}
%         \subfigure[Training loss vs. steps]{\includegraphics[width=0.23\linewidth]{figures/160m-tr-loss.png}}
%         \subfigure[KL divergence]{\includegraphics[width=0.23\linewidth]{figures/160m-kl.png}}
%     \caption{The evolution of the lower-level training loss and KL divergence for different proxy model size. Subfigures (a), (b): Proxy model size 31M, target LLM size 410M. Subfigures (c), (d): Proxy model size 160M, target LLM size 410M. }
%     \label{fig:proxy_model_size}
% \end{figure*}

\begin{table*}[!t]
    \centering
      \setlength{\tabcolsep}{1pt}
    \caption{Comparison of BLISS with different size of proxy/score model and  on zero-shot evaluation over multiple downstream datasets (410M model, 10B tokens) with 20k-step training.}
    \resizebox{1\textwidth}{!}{
    \begin{tabular}{lcccccccccc}
    \toprule
\textbf{Method} & \textbf{SciQ} & \textbf{ARC-E} & \textbf{ARC-C} & \textbf{LogiQA} & \textbf{OBQA} & \textbf{BoolQ} & \textbf{HellaSwag} & \textbf{PIQA} & \textbf{WinoGrande} & \textbf{Average} \\
  \midrule
\multicolumn{1}{l}{BLISS (Pythia-31M)} &65.5\scriptsize{(1.5)}	&40.8\scriptsize{(1.0)}	&23.4\scriptsize{(1.2)}	&27.2\scriptsize{(1.7)}	&29.8\scriptsize{(2.0)}	&58.9\scriptsize{(0.9)}	&36.0\scriptsize{(0.5)}	&67.6\scriptsize{(1.1)}	&53.4\scriptsize{(1.4)}	&44.7\scriptsize{(1.3)}\\
\multicolumn{1}{l}{BLISS (Pythia-160M)} &63.8\scriptsize{(1.5)}	&40.8\scriptsize{(1.0)}	&23.4\scriptsize{(1.2)}	&27.5\scriptsize{(1.8)}	&29.8\scriptsize{(2.0)}	&51.3\scriptsize{(0.9)}	&38.3\scriptsize{(0.5)}	&67.6\scriptsize{(1.1)}	&50.4\scriptsize{(1.4)}	&44.1\scriptsize{(1.3)}
 \\
% \multicolumn{1}{l}{BLISS (Pythia-31M without sigmoid)} &62.6\scriptsize{(1.5)}	&41.0\scriptsize{(1.0)}	&24.0\scriptsize{(1.2)}	&26.4\scriptsize{(1.7)}	&30.4\scriptsize{(2.1)}	&53.4\scriptsize{(0.9)}	&39.5\scriptsize{(0.5)}	&68.3\scriptsize{(1.1)}	&52.2\scriptsize{(1.4)}	&44.2\scriptsize{(1.3)}\\
  \bottomrule
\end{tabular}\label{tab:proxy_model_size}}
\end{table*}

\subsection{Initialization Method for the Score model} \label{sec:init_method}

In Algorithm \ref{alg:alg1}, we initialize the score model in each new round using the parameters from the last round. This design is motivated by the role of the score model: it learns data representations and ranks the importance of training samples. As training progresses, the model’s ability of feature learning improves, making it beneficial to retain learned representations across rounds.

To validate this, we conduct ablation studies comparing two cases:  
\begin{enumerate}
    \item \textbf{Original BLISS (BLISS-org)}: the score model in each round is initialized with the parameters from the last round.
    \item \textbf{Modified Initialization (BLISS$^\dag$)}: the score model in each round is reset to its initial parameters from round 1.
\end{enumerate}

We then use the trained score models from two cases to select training data and pretrain the target LLM for 15B tokens, respectively. The resulting LLMs are evaluated on multiple downstream datasets. As shown in Table \ref{tab:init_score_model}, BLISS$^\dag$ achieves an average performance that is $0.4\%$ lower than BLISS-org, demonstrating that continuous initialization leads to better data ranking and improved downstream performance.

\begin{table*}[!h]
    \centering
      \setlength{\tabcolsep}{1pt}
    \caption{Comparison of methods on zero-shot evaluation over multiple downstream datasets (410M model, 15B tokens). BLISS-org denotes the original algorithm, and BLISS$^\dag$ is a variant which uses different initialization method for the score model.}
    \resizebox{1\textwidth}{!}{
    \begin{tabular}{lcccccccccc}
        \toprule
        \textbf{Methods} (\#FLOPs $\times 10^{19}$) & \textbf{SciQ} & \textbf{ARC-E} & \textbf{ARC-C} & \textbf{LogiQA} & \textbf{OBQA} & \textbf{BoolQ} & \textbf{HellaSwag} & \textbf{PIQA} & \textbf{WinoGrande} & \textbf{Average} \\
        \midrule
        BLISS-org & 67.7 \scriptsize{(1.5)}	&41.7 \scriptsize{(1.0)}	&23.6 \scriptsize{(1.2)}	& 25.8\scriptsize{(1.7)}	& 28.4\scriptsize{(2.0)}	&56.0 \scriptsize{(0.8)}	&39.7 \scriptsize{(0.5)}	&68.7 \scriptsize{(1.1)}	&53.2 \scriptsize{(1.4)}	&44.9 \scriptsize{(1.3)} \\
        \hline
        BLISS$^\dag$   & 65.2 \scriptsize{(1.5)} & 41.6 \scriptsize{(1.0)} & 23.4 \scriptsize{(1.2)} & 27.1 \scriptsize{(1.7)} & 29.8 \scriptsize{(2.0)} & 57.5 \scriptsize{(0.8)} & 34.9 \scriptsize{(0.5)} & 67.7 \scriptsize{(1.1)} & 53.5 \scriptsize{(1.4)} & 44.5 \scriptsize{(1.3)}\\
        \bottomrule
    \end{tabular}}\label{tab:init_score_model}
\end{table*}

\subsection{Validation Datasets} \label{sec:val_data}

\begin{figure}[!t]
    \centering
    % \setlength{\tabcolsep}{1pt}
    % \resizebox{0.95\textwidth}{!}{
    \includegraphics[width=0.8\linewidth]{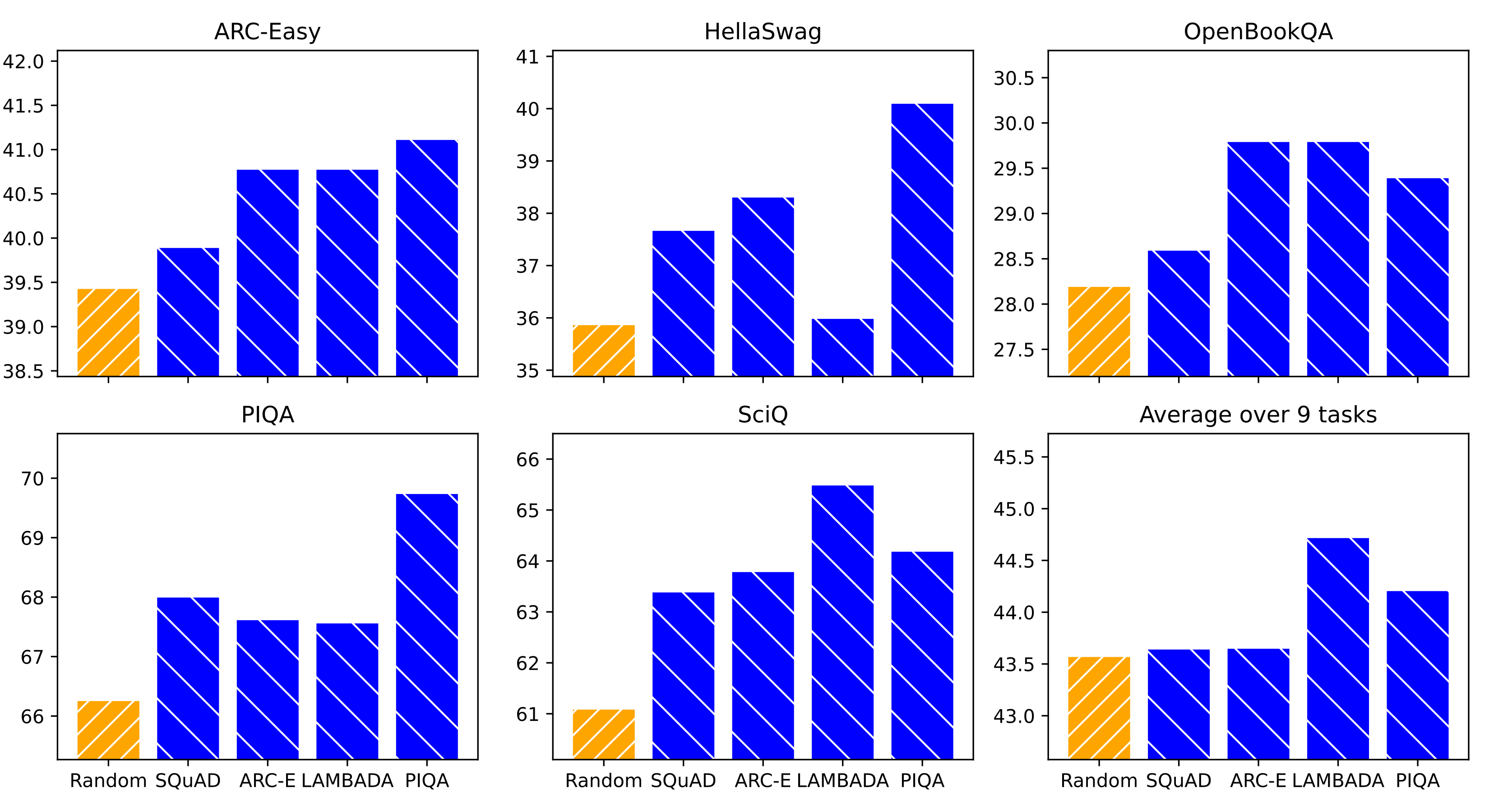} %}
    \caption{Comparison of BLISS trained with different validation datasets (410M model, 10B tokens). We compare our method with different validation datasets with random selection  on 1 downstream task in each subplot.}
    \label{fig:val}
\end{figure}

The upper-level optimization aims to minimize the proxy model’s loss on the validation dataset, meaning different validation datasets influence data selection. We use different validation set, including SQUAD, ARC-E, LAMBADA, and PIQA, to conduct the bilevel data training, then compare the corresponding downstream performance.

As shown in Figure \ref{fig:val}, our algorithm outperforms random selection on most downstream tasks, except BoolQ, regardless of the validation dataset. Notably, LAMBADA yields the highest average accuracy, improving 1.15\% over random selection, likely due to its broad domain coverage.

We additionally experimented with a mixed validation set formed by combining multiple validation datasets (SQuAD, ARC-E, LAMBADAM, PIQA), and obtained an average downstream accuracy of 44.08\% after 10B-token pretraining. When compared with the \emph{Average} result (the last subfigure) in  \cref{fig:val}, this result is lower than the best single-validation-set choice, where \textsc{LAMBADA} achieves the highest average accuracy among all tested validation sets. Therefore, our current evidence suggests that simply mixing multiple validation sets does not necessarily improve the upper-level supervision signal for BLISS. A possible reason is that \textsc{LAMBADA} used in \citep{engstrom2024dsdm,yu2024mates} provides a more coherent notion or signal of downstream utility for data selection. 

We also notice that our averaged performance is greatly affected by the accuracy of BoolQ task across all validation datasets. This indicates that it is hard to learn when the answer is too short like yes or no. 

\section{Additional results}
Since we use the same experimental settings as MATES\citep{yu2024mates}, including pretraining model, data and training steps, we evaluate MATES on the downstream tasks with their checkpoint model (\url{https://huggingface.co/yuzc19/pythia-410m-mates/blob/main/iter-200800-ckpt.pth}) of 50k  steps. For other baselines, we quote Table 1 from MATES\citep{yu2024mates} for convenience of look-up for the performance of more algorithms. 

\begin{table*}[h]
\centering
\setlength{\tabcolsep}{1pt}
\caption{ Results of Different Methods under the 410M/1B Setting. Subscripts denote standard error. Best scores are in bold.}
\resizebox{0.99\textwidth}{!}{
\begin{tabular}{l|cccccccccc}
\toprule
\textbf{Methods}$ {(\text{\#FLOPs} \times 10^{19})}$ & \textbf{SciQ} & \textbf{ARC-E} & \textbf{ARC-C} & \textbf{LogiQA} & \textbf{OBQA} & \textbf{BoolQ} & \textbf{HellaSwag} & \textbf{PIQA} & \textbf{WinoGrande} & \textbf{Average} \\
\midrule
\multicolumn{11}{l}{\textbf{410M Setting:} 410M model, 25B tokens} \\
\midrule
Random\scriptsize{(6.35)} & 64.1\scriptsize{(1.5)} & 40.2\scriptsize{(1.0)} & \textbf{25.6}\scriptsize{(1.3)} & 24.7\scriptsize{(1.7)} & 29.4\scriptsize{(2.0)} & 58.9\scriptsize{(0.9)} & 39.7\scriptsize{(0.5)} & 67.1\scriptsize{(1.1)} & 50.6\scriptsize{(1.4)} & 44.5\scriptsize{(1.3)} \\
DSIR\scriptsize{(6.35)} & 63.1\scriptsize{(1.5)} & 39.9\scriptsize{(1.0)} & 23.8\scriptsize{(1.2)} & 27.0\scriptsize{(1.7)} & 28.4\scriptsize{(2.0)} & 58.3\scriptsize{(0.9)} & 39.6\scriptsize{(0.5)} & 66.8\scriptsize{(1.1)} & 51.5\scriptsize{(1.4)} & 44.3\scriptsize{(1.3)} \\
LESS\scriptsize{(246.35)} & 64.6\scriptsize{(1.5)} & 42.3\scriptsize{(1.0)} & 23.1\scriptsize{(1.2)} & 25.2\scriptsize{(1.7)} & 30.4\scriptsize{(2.1)} & 55.6\scriptsize{(0.9)} & \textbf{41.9}\scriptsize{(0.5)} & 67.2\scriptsize{(1.1)} & 51.0\scriptsize{(1.4)} & 44.6\scriptsize{(1.4)} \\
SemDeDup\scriptsize{(7.81)} & 63.5\scriptsize{(1.5)} & \textbf{42.4}\scriptsize{(1.0)} & 24.4\scriptsize{(1.3)} & \textbf{27.6}\scriptsize{(1.7)} & 30.0\scriptsize{(2.1)} & 58.2\scriptsize{(0.9)} & 40.8\scriptsize{(0.5)} & 67.8\scriptsize{(1.1)} & 52.3\scriptsize{(1.4)} & 45.2\scriptsize{(1.3)} \\
DsDm\scriptsize{(10.72)} & {65.4}\scriptsize{(1.5)} & 41.7\scriptsize{(1.0)} & 24.7\scriptsize{(1.3)} & 27.5\scriptsize{(1.8)} & 29.0\scriptsize{(2.1)} & 57.5\scriptsize{(0.9)} & 40.3\scriptsize{(0.5)} & 67.1\scriptsize{(1.1)} & 50.1\scriptsize{(1.4)} & 44.9\scriptsize{(1.4)} \\
QuRating\scriptsize{(26.35)} & 64.8\scriptsize{(1.5)} & 42.0\scriptsize{(1.0)} & 25.4\scriptsize{(1.3)} & 25.3\scriptsize{(1.7)} & 30.2\scriptsize{(2.1)} & 58.9\scriptsize{(0.9)} & 40.7\scriptsize{(0.5)} & 67.5\scriptsize{(1.1)} & 52.1\scriptsize{(1.4)} & 45.2\scriptsize{(1.4)} \\
MATES\scriptsize{(8.11)} & 65.7\scriptsize{(1.5)} & 41.5\scriptsize{(1.0)} & 25.0\scriptsize{(1.3)} & 26.1\scriptsize{(1.7)} & \textbf{30.8}\scriptsize{(2.1)} & \textbf{60.6}\scriptsize{(0.9)} & 41.0\scriptsize{(0.5)} & 67.8\scriptsize{(1.1)} & 51.8\scriptsize{(1.4)} & 45.7\scriptsize{(1.4)} \\
BLISS\scriptsize{(8.08)}   & \textbf{68.1}\scriptsize{(1.5)} & {42.2}\scriptsize{(1.0)} & 25.1\scriptsize{(1.3)} & {27.3}\scriptsize{(1.7)} & 29.6\scriptsize{(2.0)} & 59.3\scriptsize{(0.9)} & {41.2}\scriptsize{(0.5)} & \textbf{68.2}\scriptsize{(1.1)} & \textbf{52.0}\scriptsize{(1.4)} & \textbf{45.9}\scriptsize{(1.4)}\\
\midrule
\multicolumn{11}{l}{\textbf{1B Setting:} 1B model, 25B tokens} \\
\midrule
Random\scriptsize{(17.67)} & 65.8\scriptsize{(1.5)} & 43.7\scriptsize{(1.0)} & 25.6\scriptsize{(1.3)} & 27.5\scriptsize{(1.8)} & 31.8\scriptsize{(2.1)} & 60.2\scriptsize{(0.9)} & 43.8\scriptsize{(0.5)} & 68.9\scriptsize{(1.1)} & 50.7\scriptsize{(1.4)} & 46.4\scriptsize{(1.4)} \\
DSIR\scriptsize{(17.67)} & 65.8\scriptsize{(1.5)} & 42.6\scriptsize{(1.0)} & 24.7\scriptsize{(1.3)} & \textbf{28.7}\scriptsize{(1.8)} & 29.2\scriptsize{(2.0)} & 59.7\scriptsize{(0.9)} & 44.2\scriptsize{(0.5)} & 68.3\scriptsize{(1.1)} & \textbf{53.2}\scriptsize{(1.4)} & 46.3\scriptsize{(1.4)} \\
SemDeDup\scriptsize{(19.13)} & 66.8\scriptsize{(1.5)} & 45.5\scriptsize{(1.0)} & 25.3\scriptsize{(1.3)} & 27.6\scriptsize{(1.8)} & 30.6\scriptsize{(2.1)} & 60.2\scriptsize{(0.9)} & 45.3\scriptsize{(0.5)} & 69.7\scriptsize{(1.1)} & 52.5\scriptsize{(1.4)} & 47.1\scriptsize{(1.4)} \\
DsDm\scriptsize{(22.04)} & {68.2}\scriptsize{(1.5)} & 45.0\scriptsize{(1.0)} & \textbf{26.5}\scriptsize{(1.3)} & 26.6\scriptsize{(1.7)} & 29.4\scriptsize{(2.0)} & 59.0\scriptsize{(0.9)} & 44.8\scriptsize{(0.5)} & 68.9\scriptsize{(1.1)} & 51.9\scriptsize{(1.4)} & 46.7\scriptsize{(1.3)} \\
QuRating\scriptsize{(37.67)} & 67.1\scriptsize{(1.5)} & 45.5\scriptsize{(1.0)} & 25.6\scriptsize{(1.3)} & 26.9\scriptsize{(1.7)} & 29.8\scriptsize{(2.0)} & 60.3\scriptsize{(0.9)} & 45.2\scriptsize{(0.5)} & 70.2\scriptsize{(1.1)} & 51.6\scriptsize{(1.4)} & 46.9\scriptsize{(1.3)} \\
MATES\scriptsize{(19.97)} & 67.3\scriptsize{(1.5)} & 44.9\scriptsize{(1.0)} & 25.9\scriptsize{(1.3)} & \textbf{28.7}\scriptsize{(1.8)} & 32.2\scriptsize{(2.1)} & \textbf{60.9}\scriptsize{(0.9)} & 45.3\scriptsize{(0.5)} & 69.5\scriptsize{(1.1)} & 52.4\scriptsize{(1.4)} & 47.5\scriptsize{(1.4)} \\
BLISS\scriptsize{(19.53)}   & \textbf{69.4}\scriptsize{(1.5)} & \textbf{45.7}\scriptsize{(1.0)} & 24.8\scriptsize{(1.3)} & {25.8}\scriptsize{(1.7)} & \textbf{33.2}\scriptsize{(2.1)} & 59.8\scriptsize{(0.9)} & \textbf{47.8}\scriptsize{(0.5)} & \textbf{71.6}\scriptsize{(1.1)} & 52.9\scriptsize{(1.4)} & \textbf{47.9}\scriptsize{(1.3)}\\
\bottomrule
\end{tabular}}
\label{tab:mates_results}
\end{table*}

% \begin{table*}[!h]
%     \centering
%       \setlength{\tabcolsep}{1pt}
%     \caption{Comparison of methods on zero-shot evaluation over multiuple downstream datasets (1B model, 25B tokens with 20k training steps). Best results are marked bold. The accuracy with standard error is reported based on the lm-evaluation-harness \cite{gao2021framework} implementation.}
%     \resizebox{0.99\textwidth}{!}{
%     \begin{tabular}{lcccccccccc}
%         \toprule
%         \textbf{Methods} (\#FLOPs $\times 10^{19}$) & \textbf{SciQ} & \textbf{ARC-E} & \textbf{ARC-C} & \textbf{LogiQA} & \textbf{OBQA} & \textbf{BoolQ} & \textbf{HellaSwag} & \textbf{PIQA} & \textbf{WinoGrande} & \textbf{Average} \\ 
%         \midrule
%         \multicolumn{11}{l}{\textbf{1B Setting:} 1B model, 25B tokens} \\
%         \midrule
%         Random  (6.35) & 64.1 \scriptsize{(1.5)} & 40.2 \scriptsize{(1.0)} & \textbf{25.6} \scriptsize{(1.3)} & 24.7 \scriptsize{(1.7)} & 29.4 \scriptsize{(2.0)} & 58.9 \scriptsize{(0.9)} & 39.7 \scriptsize{(0.5)} & 67.1 \scriptsize{(1.1)} & 50.6 \scriptsize{(1.4)} & 44.5 \scriptsize{(1.3)} \\
%         BLISS-1B (19.95) &63.3\scriptsize{(1.5)}	&{43.4}\scriptsize{(1.0)}	&{25.7}\scriptsize{(1.3)}	&27.8\scriptsize{(1.8)}	&29.8\scriptsize{(2.0)}	&57.1\scriptsize{(0.9)}	&41.5\scriptsize{(0.5)}	&69.3\scriptsize{(1.1)}	&51\scriptsize{(1.4)}	&45.4\scriptsize{(1.3)}\\
%         \bottomrule
%     \end{tabular}}
% \end{table*}

%%%%%%%%%%%%%%%%%%%%%%%%%%%%%%%%%%%%%%%%%%%%%%%%%%%%%%%%%%%%%%%%%%%%%%%%%%%%%%%
%%%%%%%%%%%%%%%%%%%%%%%%%%%%%%%%%%%%%%%%%%%%%%%%%%%%%%%%%%%%%%%%%%%%%%%%%%%%%%%
% \vspace{-0.2in}
\section{Experimental Hyperparameters} \label{sec:exp_setting}

\begin{table*}[!h]
    \centering
      \setlength{\tabcolsep}{1pt}
    \caption{Experimental settings.}
    \resizebox{0.95\textwidth}{!}{
\begin{tabular}{ll} 
\toprule
Hyperparameters                 & Values      \\
\midrule
\textit{Pretrain}                        &             \\
Data set                        & C4          \\
Tokens                          & 25B         \\
Model                           & Pythia-410M/1B/2.8B, LLaMA-0.5B \\
batch size                      & 512        \\
Sequence length                 & 1024        \\
Max learning rate               & 1e-3        \\
\midrule
\textit{bilevel optimization}            &             \\
Proxy/Score model               & Pythia-31M (for 410M LLM),  Pythia-160M (for 1B LLM), LLaMA-134M (for LLaMA-0.5B LLM)\\
$\gamma$                        & 1e-2        \\
$\lambda$                       & 1e-6        \\
batch size                      & 16(Pythia-410M)/32(Pythia 1B)          \\
Proxy/Score model learning rate($\eta_1 /\alpha$) & 1e-5     \\
GDLS learning rate($\eta_2$)          & 1e-2        \\
GDLS steps(\textit{K})                       & 3           \\
Score model steps               & 3k(Pythia-410M/1B)/1k(LLaMA-0.5B)          \\
Proxy model steps               & 3k(Pythia-410M/1B)/1k(LLaMA-0.5B)      \\
Initialization of score/proxy model & Randomly initialized\\
\bottomrule
\end{tabular}}
\label{tbl:hyperparam}

\end{table*}

Table \ref{tbl:hyperparam} shows the hyperparameter settings in our experiments. We use cosine learning rate scheduler in bilevel optimization, WSD\citep{yu2024mates} learning rate scheduler for pretraining and constant learning rate for GDLS. We use double loop to update the proxy model when employing 1B LLM, i.e., 5 steps for the lower level update. The experiments run on 8 A6000 GPUs with Distributed Data Parallel (DDP) strategy.
% \section{Learning Rate Schedule}
%\vspace{-0.1in}
\section{Evolution of training and validation loss} \label{loss_fig}
\begin{figure*}[t]
    \centering

    \begin{subfigure}{0.3\linewidth}
        \centering
        \includegraphics[width=\linewidth]{figures/train_loss_r1.png}
        \caption{Round 2.}
        \label{fig:bilevel_training_r2}
    \end{subfigure}
    \begin{subfigure}{0.3\linewidth}
        \centering
        \includegraphics[width=\linewidth]{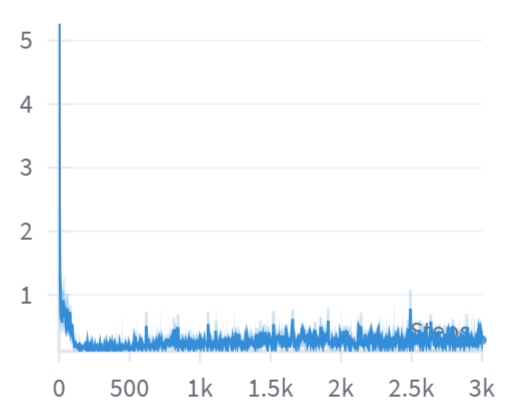}
        \caption{Round 5.}
        \label{fig:bilevel_training_r5}
    \end{subfigure}

    \caption{Evolution of training loss within a bilevel training round.}
    \label{fig:bilevel_training}
\end{figure*}

In Figure \ref{fig:bilevel_training_r2}, \ref{fig:bilevel_training_r5}, we visualize the curves training loss pretraining round 2 and 5.
% \begin{figure*}[!t]
%     \centering
%     \subfigure[ Round 2.]{\includegraphics[width=0.3\linewidth]{figures/train_loss_r1.png}
%     \label{fig:bilevel_training_r1}}\quad \quad
%     \subfigure[ Round 5.]{\includegraphics[width=0.3\linewidth]{figures/train_loss_r4.png}
%     % \subfigure[Validation loss vs. steps]{\includegraphics[width=0.46\linewidth]{figures/val_loss_r4.png}}
%     \label{fig:bilevel_training_r5}}
%     \caption{The evolution of training loss in a round.}
% \end{figure*}

% \begin{figure*}[!t]
% % \begin{minipage}{0.48\linewidth}
%     \centering
%     \subfigure[Training loss vs. steps]{\includegraphics[width=0.3\linewidth]{figures/train_loss_without_kl.png}} \quad \quad
%     \subfigure[Validation loss vs. steps]{\includegraphics[width=0.3\linewidth]{figures/val_loss_without_kl.png}}
%     \caption{The visualization of  lower-level training loss and the upper-level validation loss in round 2 bilevel optimization without KL divergence.}
%     \label{fig:training_without_kl1}
% \end{figure*}

\begin{figure*}[t]
    \centering

    \begin{subfigure}{0.3\linewidth}
        \centering
        \includegraphics[width=\linewidth]{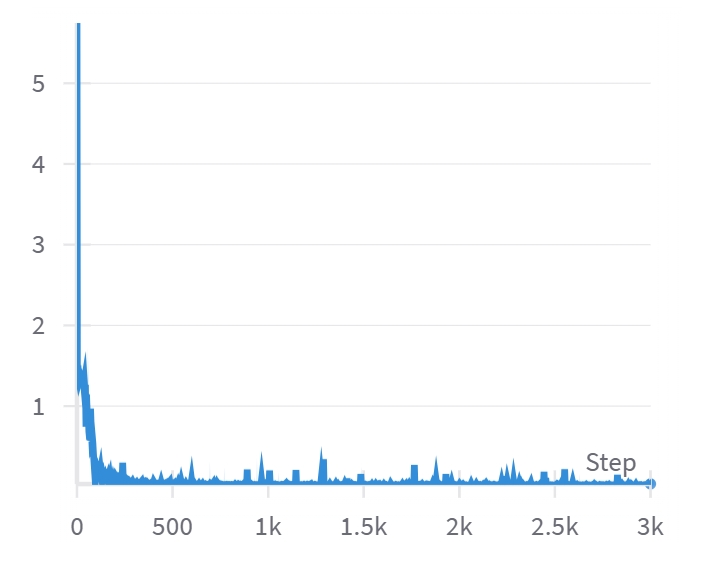}
        \caption{Training loss vs. steps.}
        \label{fig:train_without_kl}
    \end{subfigure}
    \begin{subfigure}{0.3\linewidth}
        \centering
        \includegraphics[width=\linewidth]{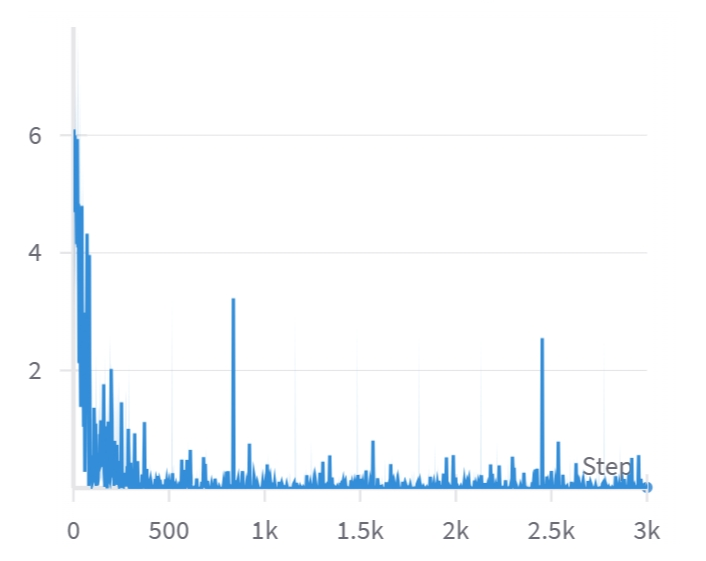}
        \caption{Validation loss vs. steps.}
        \label{fig:val_without_kl}
    \end{subfigure}

    \caption{Lower-level training loss and upper-level validation loss in round 2 of bilevel optimization without KL divergence.}
    \label{fig:training_without_kl1}
\end{figure*}

%\vspace{-0.2in}

% \section{Generalize to 1B Model}

% To assess the generality of our data selection algorithm, we extend it to a larger model, Pythia-1B. Following the experimental setup for Pythia-410M, we use the same 31M-scale proxy and score models for bilevel training. However, to accommodate the larger model, we increase the lower-level update steps (proxy model optimization) from 1 to 5 (line 9 in Algorithm \ref{alg:alg1}) and increase the batch size to 32, while keeping other hyperparameters (e.g., lower-/upper-level learning rates and bilevel training steps) unchanged. Further details on hyperparameter settings can be found in Appendix \ref{sec:exp_setting}. We report the evaluation results on downstream tasks after 20k steps of pretraining in Figure \textcolor{red}{\ref{fig:1B}}. 

\section{Distributed Softmax to Compute Influence Score}\label{sec:dist_softmax)}

\label{sec:distributedsoftmax}
In bilevel optimization, the importance weight $P_i$ is computed based on a mini batch that is distributed across different GPUs. However, back propagation through different GPUs is not implemented by Pytorch. Thus we deploy "distributed softmax" in practice. In detail, our implementation requires 3 times of communication among GPUs.  
\begin{equation}
    P_i = \frac{e^{h(\theta_s; \xi_i)}}{\sum_{j=1}^{ B} e^{h(\theta_s; \xi_j)}} \label{softmax_Pi}.
\end{equation}

As equation (\ref{softmax_Pi}) shows, the denominator of $P_i$ is the summation of every sample's exponential score. Therefore, in the first communication, each GPU gets the scores from others and calculates the denominator locally. A second communication is required to compute the term $\sum_{j=1}^\mathcal{B}  P_j \nabla_{\theta_s} h(\theta_s^t; \xi_j)$ in equation (\ref{ditributed_softmax}). In detail, we need to gather gradients of $h$ and $\mathcal{L}$' of every sample across all GPUs. After computing  hyper-gradients of every sample, they are accumulated to update upper-level variables. With efficient communication API provided by Fabric \url{https://lightning.ai/docs/fabric/stable/}, the time consumed in bilevel optimization of each round is within 1.5 hours.

\section{Running Time and Memory} \label{sec:runtime_memory}
We measured the memory and runtime of the data selection stage for both BLISS and MATES under different target (or pretraining) model sizes (for short, T: target). The results are shown in \cref{tbl:time_mem}. We have two observations:
\begin{itemize}
    \item BLISS scales well with larger target models. Note that the target model is not an optimization variable for the bilevel optimization and it is only  used  for calculating the KL divergence. Therefore, it does not affect the scalability of bilevel optimization. When increasing the target from 410M to 1B, BLISS’s memory and runtime grow moderately ($49.46\rightarrow 74.51$ GB; $5.03\rightarrow 11.82$ hours), as expected.
    \item BLISS is significantly faster than MATES. MATES incurs high cost because each round requires oracle data collection. For every example, MATES performs a one-step gradient update on the target model and evaluates the validation loss change to compute influence scores. This per-example simulation dominates runtime. In contrast, BLISS avoids all per-example oracle evaluations in MATES and performs bilevel optimization directly on the proxy/score model, leading to $2-3 \times$ faster data selection.
\end{itemize}
% These results confirm that BLISS introduces modest overhead and scales well to billion-parameter settings. 

\begin{table}[h]
\centering
\caption{The comparison of runtime and memory in data selection stage.}
\label{tbl:time_mem}
\scalebox{0.8}{
\setlength{\tabcolsep}{6pt}
\renewcommand{\arraystretch}{1.15}
\begin{tabular}{l|cc}
\toprule
\textbf{Setting} & Memory peak (GB)  & Total data selection time (hours)\\
\hline
MATES: \text{ T: 410M}         & $36.36$        & $18.0$\\
MATES: \text{ T: 1B}         & $63.52$        & $30.32$\\
\hline
BLISS: \text{T: 410M}         & $49.46$        & $5.03$ \\
BLISS: \text{T: 1B}         & $74.51$        & $11.82$ \\
\bottomrule
\end{tabular}
}% end scalebox
\end{table}

\section{Ablation Study for Bilevel Optimization Steps} \label{sec:bilevel_step}
\begin{figure*}[!t]
 \begin{minipage}{0.45\linewidth}
    \centering
    \includegraphics[width=1\linewidth]{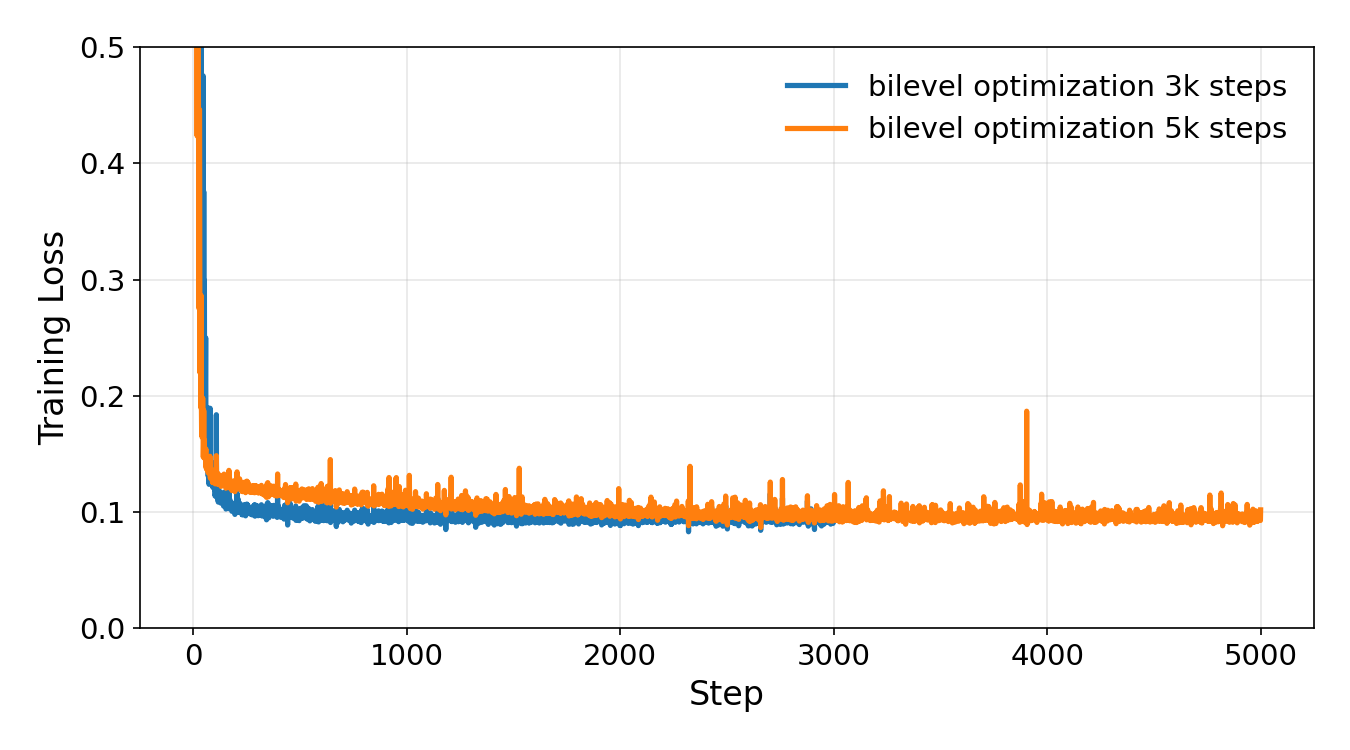}
    \caption{Training loss with different steps.}
    \label{fig:training_steps}
    \end{minipage}\quad \qquad
  \begin{minipage}{0.45\linewidth}  
      \centering
    \includegraphics[width=1\linewidth]{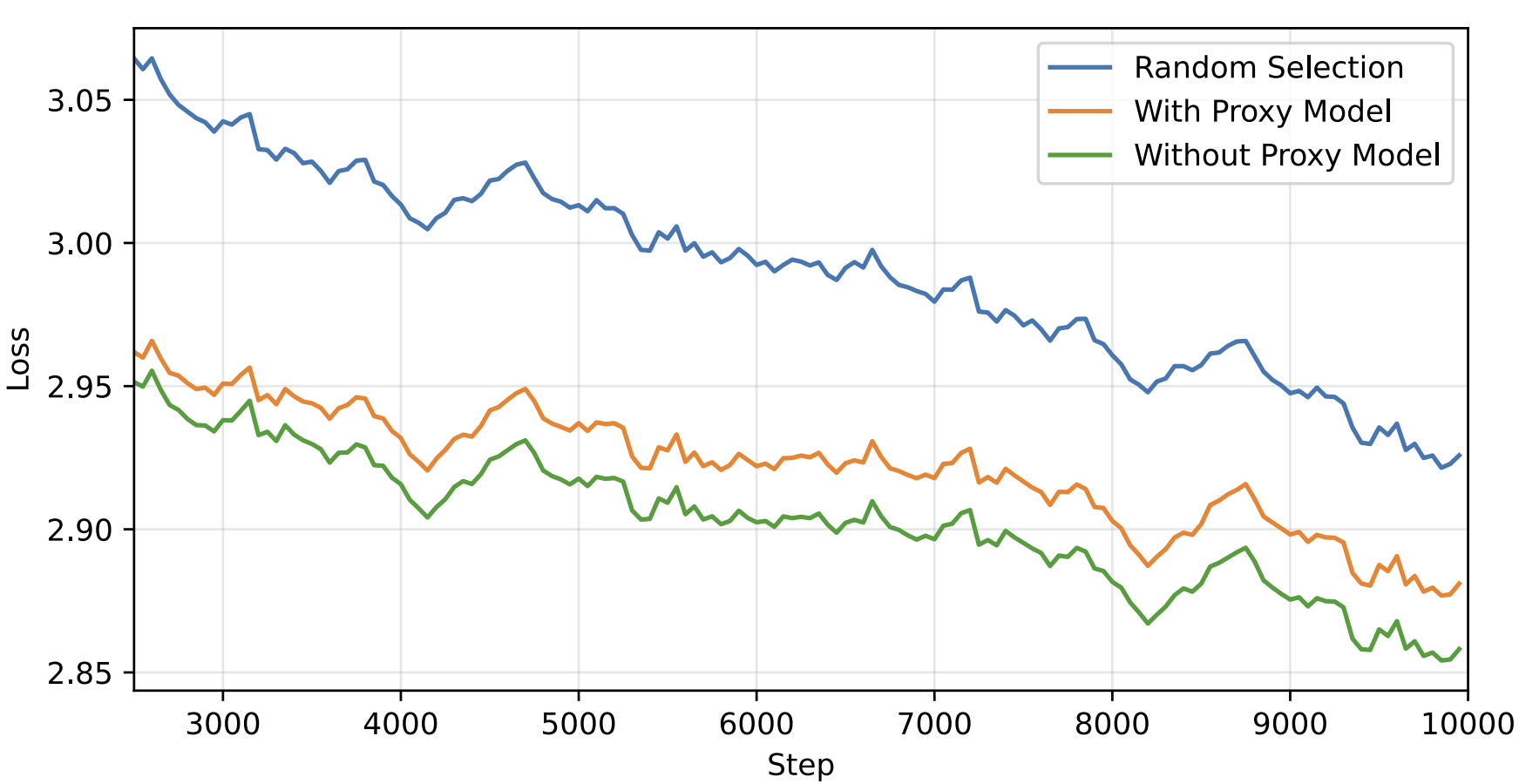}
    \caption{Test loss on SlimPajama-6B.}\label{fig:test_loss_slimpajama}
  \end{minipage}
\end{figure*}

We did the ablation study to investigate the steps of bilevel optimization on Pythia models. As shown in \cref{fig:training_steps}, both the 3k-step and 5k-step settings converge to nearly the same training loss. This indicates that 3k steps are sufficient for the Pythia-type proxy model, as increasing the steps to 5k does not yield additional improvements. So we fix the training steps of bilevel optimization to 3k steps in main experiments.

\section{Domain Reweighting: Fidelity of Proxy Models} \label{sec:domain_reweight}

To verify the fidelity of proxy models to full-scale LLMs, we conduct a domain-reweighting experiment on the SlimPajama-6B dataset~\citep{slimpajama6b}, which is a multi-domain dataset, including ArXiv, Books, C4, CommonCrawl, GitHub, StackExchange, and Wikipedia. 
The objective is to learn optimal domain weights $\alpha \in \mathbb{R}^d$, $d$ is the domain size, such that a model trained on data sampled according to the weights achieves the best downstream performance.

\begin{figure*}[!t]
    \centering

    \begin{subfigure}{0.32\linewidth}
        \centering
        \includegraphics[width=\linewidth]{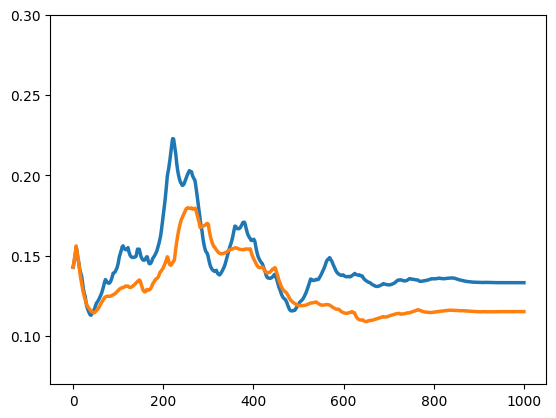}
        \caption{Wikipedia}
    \end{subfigure}
    \hfill
    \begin{subfigure}{0.32\linewidth}
        \centering
        \includegraphics[width=\linewidth]{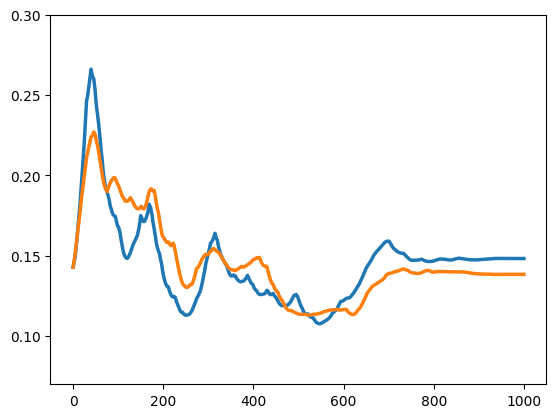}
        \caption{Book}
    \end{subfigure}
    \hfill
    \begin{subfigure}{0.32\linewidth}
        \centering
        \includegraphics[width=\linewidth]{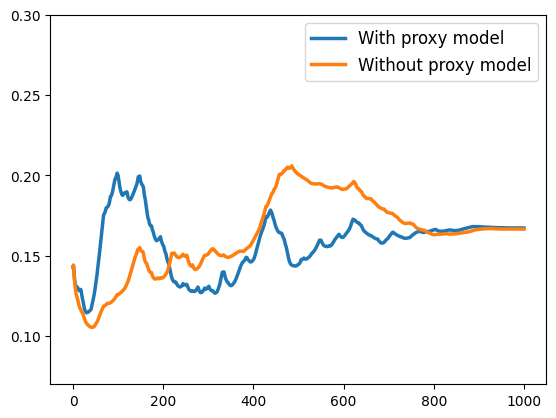}
        \caption{StackExchange}
    \end{subfigure}

    \vspace{0.5em}

    \begin{subfigure}{0.32\linewidth}
        \centering
        \includegraphics[width=\linewidth]{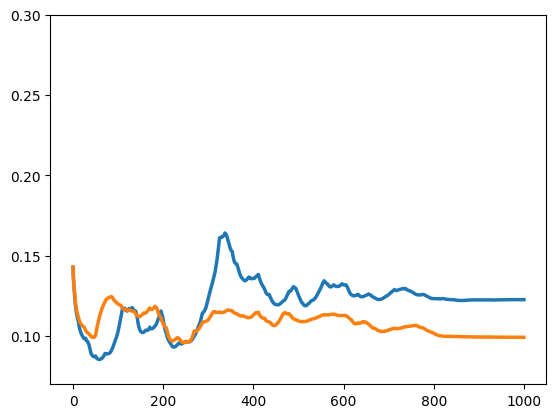}
        \caption{CommonCrawl}
    \end{subfigure}
    \hfill
    \begin{subfigure}{0.32\linewidth}
        \centering
        \includegraphics[width=\linewidth]{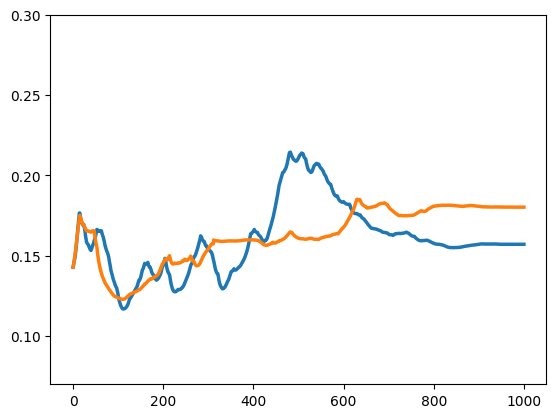}
        \caption{ArXiv}
    \end{subfigure}
    \hfill
    \begin{subfigure}{0.32\linewidth}
        \centering
        \includegraphics[width=\linewidth]{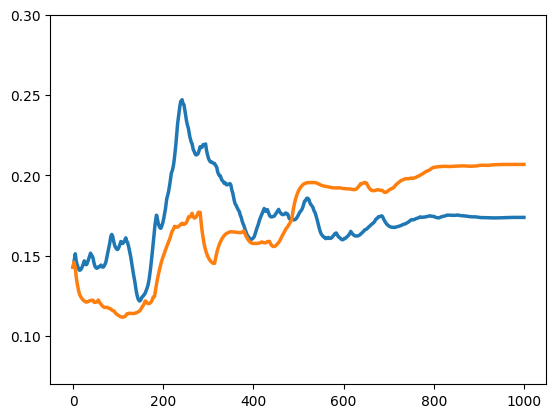}
        \caption{GitHub}
    \end{subfigure}

    \caption{Learning curves of domain weights.}
    \label{fig:domain_weight_cruves}
\end{figure*}

We compare two settings:
\begin{enumerate}
    \item \textbf{Case 1 (with proxy model):}  
    The lower level optimizes a lightweight proxy model (LLaMA-134M) with output alignment to the target LLM (LLaMA-300M), and the upper level learns the domain weights $\tilde{\alpha}$.

    \item \textbf{Case 2 (without proxy model):}  
    The lower level directly optimizes the target LLM (LLaMA-300M), and the upper level learns the domain weights $\alpha$.
\end{enumerate}
We perform bilevel optimization for 1,000 steps in both cases to learn the domain weights, where $10\%$ of the original training set is held out as the validation set for the upper-level objective, and the remaining $90\%$ is used as the lower-level training set.
After obtaining $\tilde{\alpha}$ and $\alpha$, we train two final LLaMA-300M models on data sampled according to each set of weights, respectively . 
Figure \ref{fig:domain_weight_cruves} presents the learning curves of domain weights for both cases. 
We observe that the trajectories of $\tilde{\alpha}$ and $\alpha$ are highly similar across most domains (e.g., in the domain of Wikipedia, Book, Stackexchange), demonstrating that the proxy model maintains high fidelity to the full-scale LLM in data selection.

Finally, we evaluate the resulting pretrained LLMs on the test set of SlimPajama-6B, and the results are shown in Figure~\ref{fig:test_loss_slimpajama}. The test loss curves show that data selection based on the proxy model maintains high fidelity to the full-scale LLM, while significantly outperforming random selection.

% \begin{figure*}[!t]
%     \centering
%     \subfigure[Wikipedia]{\includegraphics[width=0.3\linewidth]{figures/wikipedia.png}}
%     \subfigure[Book]{\includegraphics[width=0.3\linewidth]{figures/book.png}}
%     \subfigure[Stackexchange]{\includegraphics[width=0.3\linewidth]{figures/stackexchange.png}}
%     \subfigure[Commoncrawl]{\includegraphics[width=0.3\linewidth]{figures/commoncrawl.png}}
%     \subfigure[Arxiv]{\includegraphics[width=0.3\linewidth]{figures/arxiv.png}}
%     \subfigure[Github]{\includegraphics[width=0.3\linewidth]{figures/github.png}}
%     \caption{Learning curves of domain weights.}
%     \label{fig:domain_weight_cruves}
% \end{figure*}

%%%%%%%%%%%%%%%%%%%%%%%%%%%%%%%%%%%%%%%%%%%%%%%%%%%%%%%%%%%%%%%%%%%%%%%%%%%%%%%
%%%%%%%%%%%%%%%%%%%%%%%%%%%%%%%%%%%%%%%%%%%%%%%%%%%%%%%%%%%%%%%%%%%%%%%%%%%%%%%

\end{document}